\documentclass[10pt,twocolumn,letterpaper]{article}
\usepackage{cvpr}
\usepackage{times}
\usepackage{epsfig}
\usepackage{graphicx}
\usepackage{amsmath}
\usepackage{amssymb}

\usepackage[pagebackref=true,breaklinks=true,letterpaper=true,colorlinks,bookmarks=false]{hyperref}
\usepackage{booktabs}
\usepackage{subfigure}
\usepackage{multirow}
\cvprfinalcopy % *** Uncomment this line for the final submission

\def\cvprPaperID{3387} % *** Enter the CVPR Paper ID here

\ifcvprfinal\fi

\def\FL{\textrm{FL}}
\def\pt{p_\textrm{t}}
\usepackage{authblk}
\newcommand{\eqnnm}[2]{\begin{equation}\label{eq:#1}#2\end{equation}\ignorespaces}

\begin{document}

%%%%%%%%% TITLE
\title{Strong-Weak Distribution Alignment for Adaptive Object Detection
}
\author{Kuniaki Saito$^1$ Yoshitaka Ushiku$^2$ Tatsuya Harada$^{2,3}$ Kate Saenko$^1$ \\$^1$Boston University, $^2$The University of Tokyo, $^3$RIKEN\\
\tt\small $^1$\{keisaito, saenko\}@bu.edu, $^2$\{ushiku, harada\}@mi.t.u-tokyo.ac.jp}

\maketitle

%\author{Kuniaki  Saito\\
%Boston University\\
%{\tt\small keisaito@bu.edu}
%\and
%Yoshitaka Ushiku\\
%The University of Tokyo\\
%{\tt\small ushiku@mi.t.u-tokyo.ac.jp}
%\and
%Tatsuya Harada\\
%The University of Tokyo\\
%{\tt\small harada@mi.t.u-tokyo.ac.jp}
%\and
%Kate  Saenko\\
%Boston University\\
%{\tt\small saenko@bu.edu}
%}

\maketitle
%\thispagestyle{empty}

%%%%%%%%% ABSTRACT
\begin{abstract}

We propose an approach for unsupervised adaptation of object detectors from label-rich to label-poor domains which can significantly reduce annotation costs associated with detection. Recently, approaches that align distributions of source and target images using an adversarial loss have been proven effective for adapting object classifiers. However, for object detection, fully matching the entire distributions of source and target images to each other at the global image level may fail, 
as domains could have distinct scene layouts and different combinations of objects. On the other hand, strong matching of local features such as texture and color makes sense, as it does not change category level semantics. This motivates us to propose a novel method for detector adaptation based on strong local alignment and weak global alignment. Our key contribution is the weak alignment model, which focuses the adversarial alignment loss on images that are globally similar and puts less emphasis on aligning images that are globally dissimilar. Additionally, we design the strong domain alignment model to only look at local receptive fields of the feature map. We empirically verify the effectiveness of our method on four datasets comprising both large and small domain shifts. Our code is available at \url{https://github.com/VisionLearningGroup/DA_Detection}.

\end{abstract}

%%%%%%%%% BODY TEXT
\section{Introduction}
Deep convolutional neural networks have greatly improved object recognition accuracy~\cite{krizhevsky2012imagenet}, but remain reliant on large quantities of labeled training data. For object detection, annotation is particularly burdensome: each instance of an object category in every image must be annotated with a precise bounding box. Transferring pre-trained models from label-rich domains is an attractive solution, but dataset bias often reduces their generalization to novel data~\cite{saenko2010}. 

\begin{figure}
    \centering
    \includegraphics[width=0.95\hsize]{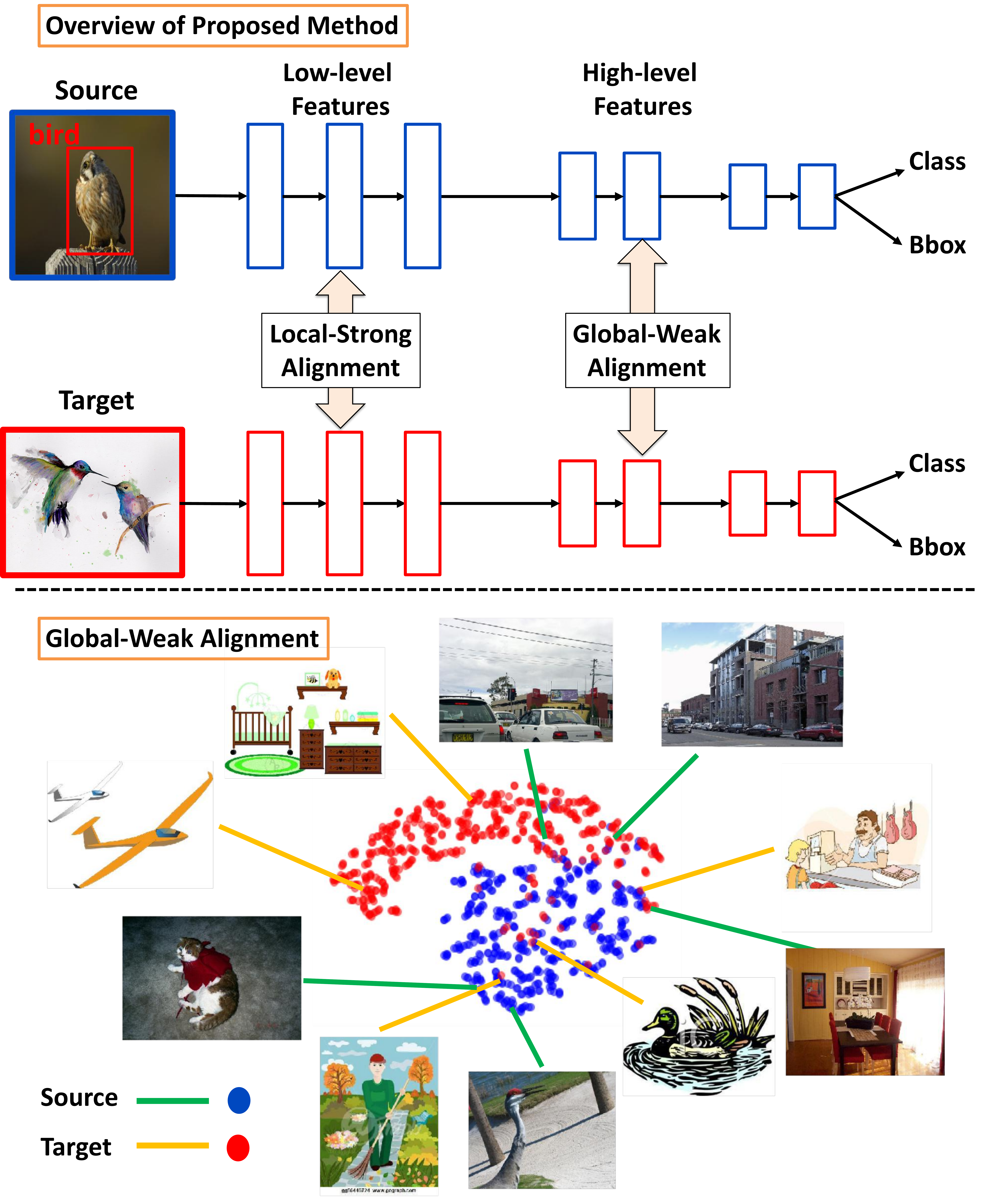}
    \caption{\small Upper: 
    Our Strong-Weak model learns domain-invariant features that are strongly aligned at the local patch level and weakly (partially) aligned at the global scene level.
    Lower: Global features obtained by our proposed weak alignment method on Pascal to Clipart. The target features are partially aligned with source, which improves detection performance, as shown in our experiments.}
    \label{fig:fig1}
\end{figure}

Various methods for unsupervised domain adaptation (UDA) have been proposed to tackle the dataset bias problem~\cite{ganin2016domain,tzeng2014deep,tzeng2017adversarial,long2016unsupervised}, most of which are based on domain-invariant alignment of the feature \cite{ADR} or image~\cite{unit,cycada} distributions. 
Recent methods align the source and target distributions of examples using adversarial learning and are motivated by theoretical results that bound the generalization error partially by the size of the discrepancy between domains~\cite{ben2007analysis,ben2010theory}. The conventional wisdom is therefore that discrepancy must be reduced at all costs, which can only be done if one fully aligns the distributions. In this paper, we argue that such \textit{strong} domain alignment is only reasonable in closed problems, such as object classification settings where the source and target examples share the same categories and prior label distributions. In settings such as open-set classification~\cite{busto2017open,saito2018open} or partial domain adaptation \cite{zhang2018importance}, strong alignment can be infeasible and could actually hurt performance. 

In object detection this is particularly evident, as aligning global (image-level) features means that not only the object categories, but also backgrounds and scene layouts must be similar across domains. Yet this is precisely what the current state-of-the-art UDA method for detection, Adaptive Faster RCNN~\cite{dafaster}, attempts to do. It trains Faster RCNN with a domain classifier trained to distinguish source and target examples, while the feature extractor learns to deceive the domain classifier. Feature alignment is done both at the global image scale and at the instance (object) scale.

While the global matching might work well for small domain shifts that only affect the appearance/texture of objects (e.g. weather related shifts), it is likely to hurt performance for larger shifts that affect the layout of the scene, the number of objects and/or their co-occurrence. For example, source images may contain single objects, while target images may contain multiple smaller objects. Forcing invariance to such \textit{global} features can hurt performance.%remove information required for object detection. 
On the other hand, strong alignment of \textit{local} features would match the texture or color of the domains and should improve performance in most cases, because it will not change the category information but is likely to reduce the domain gap. In this paper, by ``local" scale we do not mean the instance (object) scale but rather texture or color features with small receptive fields. 

Motivated by these observations, we propose an unsupervised adaptation method for object detection that combines \textit{weak global} alignment with \textit{strong local} alignment, called the Strong-Weak Domain Alignment model (top of Fig.~\ref{fig:fig1}). We propose to apply \textit{weak} alignment to the \textit{global} features, partially aligning them to reduce the domain gap without hurting the performance of the model. We show an example of weak global alignment in the bottom of Fig.~\ref{fig:fig1}, where only the target images which contain one object are aligned with the source. 
Our key contribution is the weak global alignment model, which focuses the adversarial alignment loss toward images that are globally similar, and away from images that are globally dissimilar. 
Additionally, we achieve  strong local alignment by constructing a domain classifier
designed to look only at local features and to strictly align them with the other domain.
We verify the effectiveness of our method in adaptation between both similar and dissimilar domains. 
\section{Related Work}
\vspace{-1mm}
%\subsection
\noindent \textbf{Object Detection.}
The development of deep convolutional neural networks has boosted the performance of object detection. Having a strong backbone feature extractor is key for accurate detection models.  Current detection networks can be categorized into two types: two-stage  and one-stage. Faster-RCNN (FRCNN) \cite{faster} is a representative two-stage detector that generates coarse object proposals using region proposal networks (RPN) as the first stage, and feeds the proposals and cropped features into a classification module as the second stage. In this paper, we use the FRCNN as a base detector, however, our method should be applicable to  other two-stage detectors and one-stage detectors such as YOLO \cite{yolo} or SSD \cite{ssd}. 
Detector back-bone networks are usually pre-trained on ImageNet~\cite{imagenet} and need to be fine-tuned again with a large number of annotated object bounding boxes. Various datasets have been publicized for this purpose \cite{everingham2010pascal, imagenet, coco}. To deal with the deficit in such large annotated datasets, weakly supervised and semi-supervised object detection has been proposed in the literature~\cite{tang2016large, bilen2016weakly}. Although cross-domain object detection and especially unsupervised cross-domain object detection can also help with this problem, as far as we know, there is only one work that has tackled the task of unsupervised domain transfer of deep object detectors~\cite{dafaster}. In this work, the feature alignment at the instance (object) scale was done for features cropped by region proposals. To effectively conduct feature alignment, region proposals have to precisely localize objects of interest. However, this is difficult to do for the target domain as we are not given ground truth proposals. The feature alignment may therefore hurt the performance of the model as we show in our experiments, which is why we do not conduct instance scale alignment in our work. 
%We did not conduct the instance scale alignment because instance scale alignment was 

%\subsection
\noindent \textbf{Domain Adaptation.}
The problem of bridging a gap between domains has been investigated for various visual applications such as image classification and semantic segmentation~\cite{saenko2010,tzeng2014deep,udaself,learnfromsynth}.
%The degradation in performance when testing a model in a different domain is mainly due to a difference in feature distributions. 
%Therefore, to
To solve the problem, a large number of methods utilize feature distribution matching between training and testing domains. 
The basic idea is to measure some type of distance between different domains' feature distributions and train a feature extractor to minimize that distance. 
Various ways of measuring the distance have been proposed \cite{ganin2014unsupervised,tzeng2014deep,tzeng2017adversarial,long2015learning,long2016unsupervised, saito2017maximum}. 
Motivated by a theoretical result~\cite{ben2007analysis,ben2010theory}, various approaches utilize the domain classifier~\cite{ganin2014unsupervised,tzeng2014deep,tzeng2017adversarial} to measure domain discrepancy. They train a domain classifier and feature extractor in an adversarial way, as done for training GANs~\cite{GAN}. Such methods are  designed to strictly align the feature distribution of the target with that of the source. In addition, Long \etal designed a loss function of the domain classifier to fully match features between domains \cite{long2017conditional} for image classification. %Strictly matching feature distributions should be helpful in a standard classification because images in the target domain necessarily belong to a certain class of the source and label-distribution should be similar.  

In this paper, we instead propose a weak feature alignment model for global features, and use strong alignment only at the local level to strictly align the style of images across domains.
Some research on GANs and domain adaptive semantic segmentation has shown that regularizing the domain classifier with task-specific classification loss can stabilize the adversarial training~\cite{acgan,learnfromsynth}. Motivated by this approach, we further propose a method to regularize the domain classifier by the detection loss on source examples.

\section{Method}
\vspace{-2mm}
The architecture of our proposed Strong-Weak DA model is illustrated in Fig.~\ref{fig:architecture}. We 
%pick up
extract  global features just before the RPN and local features from lower layers,
%\ks{Explain where the features come from, i.e. the top part of the figure.}
and perform \textit{weak global} alignment  in the high-level feature space and \textit{strong local} alignment in the low-level feature space. We further propose to stabilize the training of domain classifiers with the detection loss (Sec.~\ref{sec:context}).
%by using context vector.

%\ks{can we expand on this somehow? it needs more motivation/explanation..was it ever done before? if yes then mention it in related work?} In this section, we will explain each module.

%In our experiments, we used the Faster CNN \cite{faster} as a backbone network. However, we think our method is applicable to other kinds of detectors. 
\vspace{-2mm}
\subsection{Weak Global Feature Alignment}
\begin{figure}
    \centering
    \includegraphics[width=1.0\linewidth]{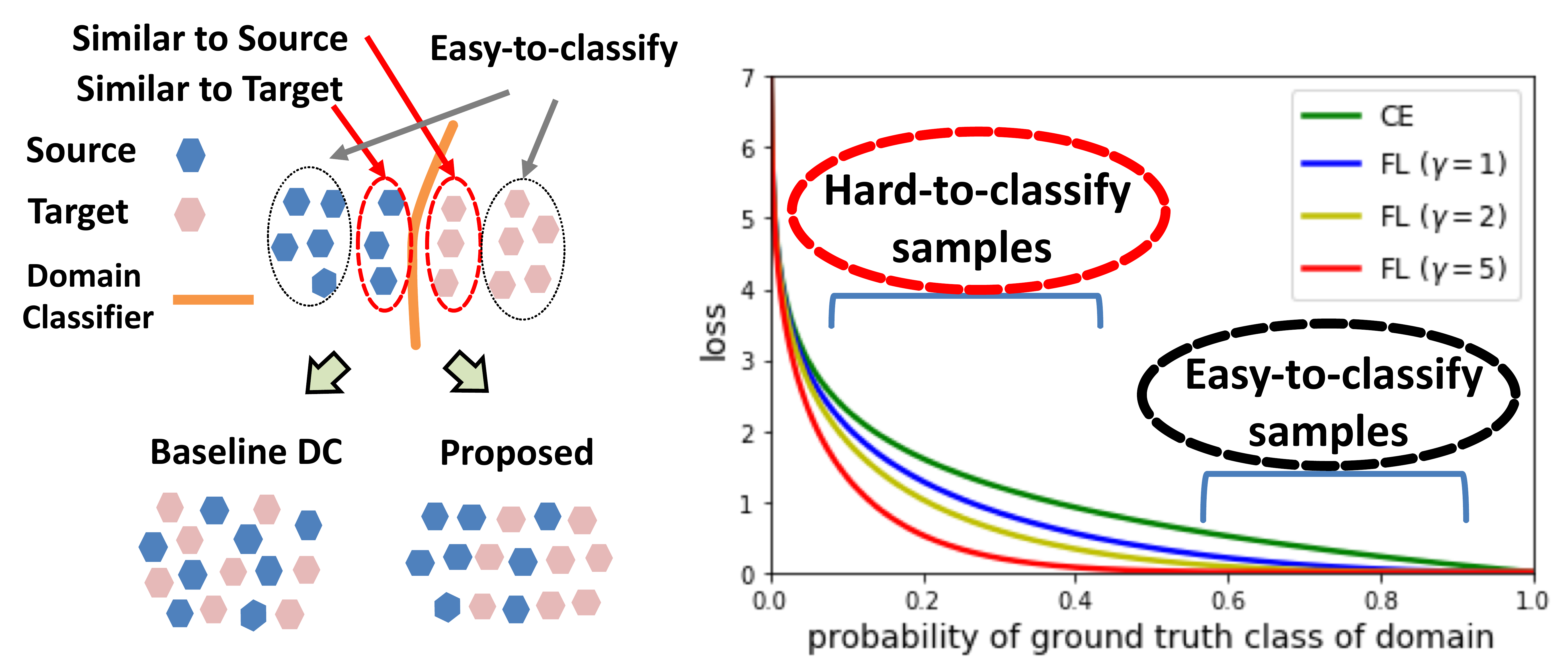}
    \caption{Left: Weak-distribution alignment using a domain classifier. Right: Standard cross-entropy loss and focal loss. }
    \label{fig:focalloss}
\end{figure}

\begin{figure*}
    \centering
    \includegraphics[width=1.0\linewidth]{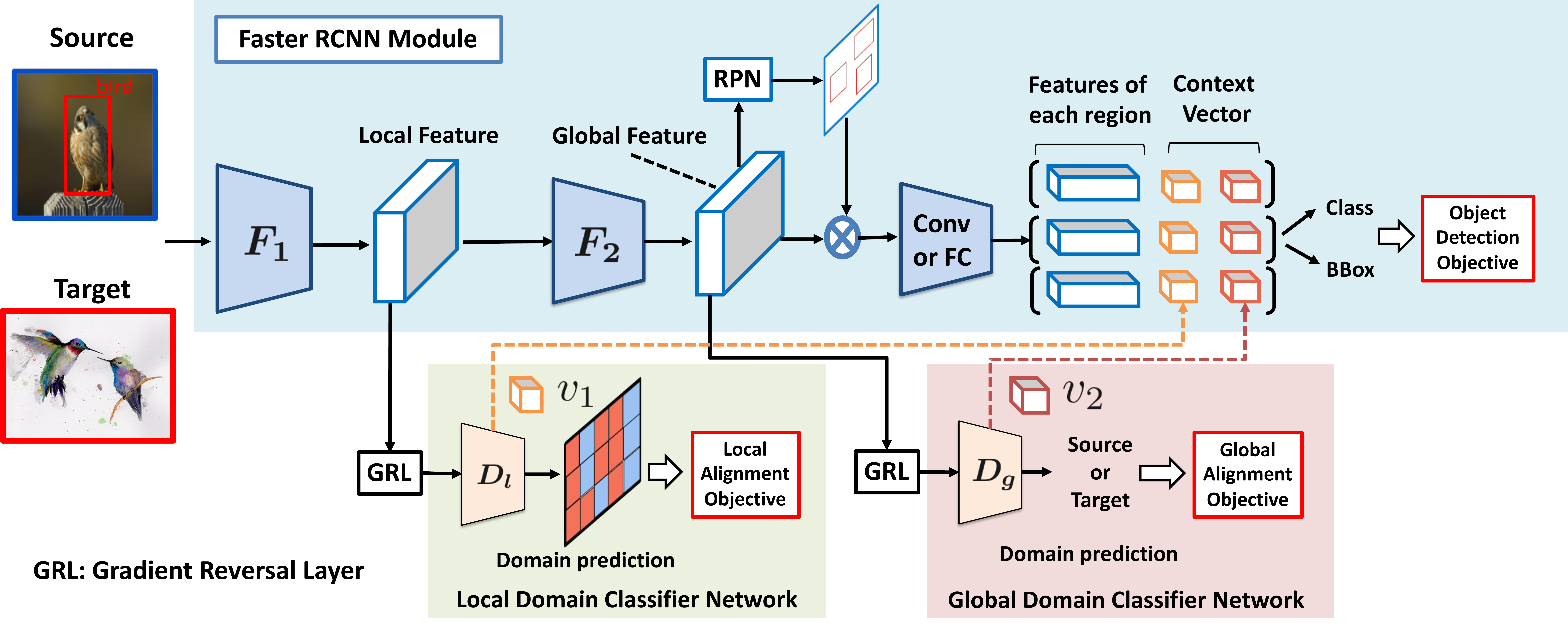}
    \caption{Proposed Network Architecture. Our method performs strong-local alignment by a local domain classifier network and weak-global alignment by a global domain classifier. The context vector is extracted by the domain classifiers and is concatenated in the layer before the final fully connected layer.} %\ks{The figure is more clear with the shading of modules at the bottom. Also shade the Faster RCNN module?} }
    \label{fig:architecture}
\end{figure*}

\vspace{-2mm}
We utilize a domain classifier to align the target features with the source for the global-level feature alignment. 
Easy-to-classify target examples are far from source examples in the feature space while hard-to-classify target examples are near the source as shown in the left of Fig. \ref{fig:focalloss}. Therefore, focusing on hard-to-classify examples should achieve a weak alignment between domains. 
We propose to train a domain classifier to ignore easy-to-classify examples while focusing on hard-to-classify examples with respect to the classification of the domain. 
%First, we will show a baseline domain classifier based method, and present our proposed method. 
%\noindent
%\textbf{Baseline domain classifier based method}

We have access to a labeled source image $x^{s}$ and bounding boxes for each image $y^{s}$ drawn from a set of annotated source images \{$X_{s},Y_{s}$\},
as well as an unlabeled target image $x^{t}$ drawn from unlabeled target images $X_{t}$.
The global feature vector is extracted by $F$.
The domain classifier, $D_g$, is trained to predict the domain of input global features.
Our learning formulation optimizes $F$ so that the features are discriminative for the primary task of object detection,
but are uninformative for the task of domain classification. The domain-label $d$ is $1$ for the source and $0$ for the target. 
The network $R$ takes features from $F$ and outputs bounding boxes with a class label. $R$ includes the Region Proposal Network (RPN) and other modules in Faster RCNN.
The objective of the detection loss is summarized as: 
\vspace{-3mm}
\begin{align}
 \mathcal{L}_{cls} (F,R) =  -\frac{1}{n_s}\sum_{i=1}^{n_s}  \mathcal{L}_{det} (R(F({x_i}^s)), {y_{i}}^{s})
\end{align}
where we assume that $ \mathcal{L}_{det}$ contains all losses for detection such as a classification loss and a bounding-box regression loss. ${n_s}$ denotes the number of source examples.

In existing methods~\cite{dafaster}, the objective for  domain classification is the cross-entropy loss. As shown in Fig. \ref{fig:focalloss},  the loss of the easy-to-classify examples, which have high probability, is not negligible in this cross-entropy loss. This indicates that $D_g$ and $F$ account for all examples in the training procedure. Therefore, $F$ tries to match the entire feature distribution, which is not desirable in domain adaptive object detection. 

%In the baseline method, the cross-entropy (CE) loss for binary classification is used to train the domain classifier. The domain-label $y_d$ is zero for the source and one for the target. 
%The loss function of the domain classification $ \mathcal{L}_{global}$ is defined as follows:
%\begin{align*}
% \mathcal{L}_{global_{s}} (F,D_g) = -\frac{1}{n_s}\sum_{i=1}^{n_s} \log(D_{g}(F({x_i}^s)))\\
% \mathcal{L}_{global_{t}} (F,D_g) = -\frac{1}{n_t}\sum_{i=1}^{n_t} \log(1 -D_{g}(F({x_i}^t)))
% \end{align*}
% \begin{align*}
%  \mathcal{L}_{global} (F,D_{g}) = \frac{1}{2}(  \mathcal{L}_{global_{s}} +  \mathcal{L}_{global_{t}} ) 
%  \end{align*}
% where $n_s$ and $n_t$ indicate number of examples in the source and target respectively.

%The overall objective is, 
%$$ \max_{D_g}\min_{F,R}  \mathcal{L}_{cls} (F,R) - \lambda  \mathcal{L}_{global} (F,D_g) %$$
%where $ \lambda$ is the trade-off parameter between two objectives. 

%In this adversarial learning, the objective used for the domain classification is the cross-entropy loss. As shown in Fig. \ref{fig:focalloss}, since the loss of the easy-classified examples, which have high probability,  is not negligible in this cross-entropy loss. This indicates that $D_g$ and $F_g$ account for all examples in this training procedure. Therefore, $F_g$ tries to match the entire feature distributions, which is not desirable in domain adaptive object detection.
%\noindent
%\textbf{Weak-global alignment}

Instead,
%Then, we will introduce how to
we want the domain classifier to ignore easy-to-classify examples while focusing on hard-to-classify examples.
%with respect to the classification of the domain. 
The problem with cross-entropy (CE) loss ($-\log p$) is that it puts non-negligible values of easy-to-classify examples where $p \in [0, 1]$ is the model's estimated probability for the class with label $d = 1$.
We propose to add a modulating factor $f(\pt)$ to the cross-entropy loss, resulting in 
\begin{equation}\label{eq:decrease}
    - f(\pt) \log (\pt)
\end{equation} 
where we define $\pt$:
 \eqnnm{pt}{\pt=\begin{cases} p &\text{if $d = 1$}\\ 1 - p &\text{otherwise.}\end{cases}} We choose a function  that decreases as $\pt$ increases. One example of such a loss function is Focal Loss (FL) \cite{focal}
 \eqnnm{fl}{\FL(\pt) = -f(\pt) \log (\pt), f(\pt) =  (1 - \pt)^\gamma}
  where $\gamma$ controls the weight on hard-to-classify examples.
FL is designed to put more weight on hard-to-classify examples than on easy ones during training, as shown in the right of Fig.~\ref{fig:focalloss}. 
% The loss of easy-to-classify samples will become negligible. The domain classifier trained with the loss will ignore easy-to-classify samples.
%The domain classifier trained with the such loss function will ignore the easy-to-classify examples. 
The feature extractor tries to deceive the domain classifier, that is, tries to increase the loss. However, the feature extractor cannot align the well-classified target examples with the source because the scale of gradients of such examples is very small. The same  can be said about aligning source examples to the target.
$f(\pt)$ can take other formulations if it satisfies the requirement described above. In experiments, we will show the result of another loss function that satisfies the condition.
We denote the loss of the weak global-level domain classifier as $ \mathcal{L}_{global}$ as follows, 
\begin{eqnarray}\label{eq:fl_t}
\scalebox{0.9}{$\displaystyle \mathcal{L}_{global_s} =  -\frac{1}{n_s}\sum_{i=1}^{n_s} (1-D_{g}(F({x_i}^s))^\gamma \log(D_{g}(F({x_i}^s))) $}
\end{eqnarray}
\vspace{-2mm}
\begin{eqnarray}\label{eq:fl_s}
\scalebox{0.9}{$\displaystyle  \mathcal{L}_{global_t}=-\frac{1}{n_t}\sum_{i=1}^{n_t}D_{g}(F({x_i}^t))^\gamma \log(1 - D_{g}(F({x_i}^t)))  $}
\end{eqnarray}
\vspace{-2mm}
\begin{align}
 \mathcal{L}_{global} (F,D_{g}) = \frac{1}{2}(  \mathcal{L}_{global_s} +  \mathcal{L}_{global_t} ) 
 \end{align}
 where ${n_t}$ denotes the number of target examples.
 
The gradients of this loss should change the parameters of low-level layers, which should also align low-level features, but the effect may not be strong enough.
%with respect to category level semantics. Since the alignment of texture and color will not be enough, 
We thus propose to directly perform the alignment in local-level features in the next sub-section. 

\subsection{Strong Local Feature Alignment}
\vspace{-2mm}
%As the goal is local-level alignment, 
The architecture of the local domain classifier, $D_{l}$, is designed to focus on the local features rather than global features. $D_{l}$ is a fully-convolutional network with kernel-size equal to one.
%The architecture of the domain classifier is, $Conv-ReLU-Conv-ReLU-Conv-Sigmoid$. 
%Kernel-size of the convolution is set one so that the domain classifier looks at only the local receptive field. 
The feature extractor $F$ is decomposed as $F_2 \circ F_1$ and the output of $F_1$ is the input to $D_{l}$ as shown in Fig. \ref{fig:architecture}. $F_1$ outputs a feature whose width and height is $W$ and $H$ respectively. $D_{l}$ outputs a domain prediction map which has the same width and height as the input feature. We employed a least-squares loss to train the domain classifier  following~\cite{least,zhu2017unpaired}. This loss function stabilizes the training of the domain classifier and is empirically shown to be useful for aligning low-level features. 
The loss function of the strong local alignment $ \mathcal{L}_{loc}$ is summarized as
\begin{align}
  \mathcal{L}_{loc_s} = \frac{1}{n_s H W}\sum_{i=1}^{n_s}\sum_{w=1}^{W}\sum_{h=1}^{H} D_{l}(F_{1}({x_i}^s))_{wh}^2\\
  \mathcal{L}_{loc_t} = \frac{1}{n_t H W}\sum_{i=1}^{n_t}\sum_{w=1}^{W}\sum_{h=1}^{H} (1 - D_{l}(F_{1}({x_i}^t))_{wh})^2
\end{align}
\begin{align}
 \mathcal{L}_{loc} (F,D_{l}) = \frac{1}{2}(  \mathcal{L}_{loc_s} +  \mathcal{L}_{loc_t} ) 
 \end{align}
  where $D_{l}(F_{1}({x_i}^s))_{wh}$ denotes the output of the domain classifier in each location.
The loss is designed to align each receptive field of features with the other domain. 

\begin{table*}[h]
\caption{Results on adpatation from PASCAL VOC to Clipart Dataset. Average precision (\%) is evaluated on target images. G, I, CTX, L indicate global alignment, instance-level alignment, context-vector based regularization, and local-alignment respectively.}
\label{tbl:ap_clipart}
  \centering
\scalebox{0.85}{
  \tabcolsep=1.5pt
  \centering
\begin{tabular}{c|cccc|ccccccccccccccccccccc}
\toprule[1.5pt]
%  & \multicolumn{20}{c}{AP for each class} & \\
%                \cmidrule(r){2-21}  
Method & G & I & CTX &L& aero & bcycle & bird & boat & bottle & bus  & car  & cat  & chair & cow  & table & dog  & hrs & bike & prsn & plnt & sheep & sofa & train & tv & MAP  \\\hline
Source Only&&&&&\bf{35.6}      & 52.5    & 24.3 & 23.0 & 20.0   & 43.9 & 32.8 & 10.7 & 30.6  & 11.7 & 13.8        & 6.0  & \bf{36.8}  & 45.9      & 48.7   & 41.9        &\bf{16.5}  & 7.3  & 22.9  & 32.0      & 27.8 \\
BDC-Faster &$\checkmark$&&&&20.2      & 46.4    & 20.4 & 19.3 & 18.7   & 41.3 & 26.5 & 6.4  & 33.2  & 11.7 &\bf{ 26.0}        & 1.7  & 36.6  & 41.5      & 37.7  & 44.5        & 10.6  & 20.4 & 33.3  & 15.5      & 25.6 \\
DA-Faster&$\checkmark$&$\checkmark$&&&15.0&34.6&12.4&11.9&19.8&21.1&23.2&3.1&22.1&26.3&10.6&10.0&19.6&39.4&34.6&29.3&1.0&17.1&19.7&24.8&19.8
\\\hline
\multirow{3}{*}{Proposed} &$\checkmark$&&&&30.5      & 48.5    & \bf{33.6} & 24.8 & 41.2   & 48.9 & 32.4 & 17.2 & 34.5  & 55.0 & 19.0        & 13.6 & 35.1  & 66.2      & \bf{63.0}   & 45.3        & 12.5  & 22.6 & 45.0  & 38.9      & 36.4 \\
 &$\checkmark$&&$\checkmark$&&31.7      &\bf{55.2}    & 30.9 & 26.8 & \bf{43.4}   & 47.5 & \bf{40.0} & 7.9  &\bf{36.7}  & 50.0 & 14.3        & \bf{18.0} & 29.2  & \bf{68.1}      & 62.3   & 50.4        & 13.4  & 24.5 & \bf{54.2}  & 45.8      & 37.5\\ 
 &$\checkmark$&&$\checkmark$&$\checkmark$&26.2&48.5&32.6&\bf{33.7}&38.5&\bf{54.3}&	37.1&\bf{18.6}&34.8&\bf{58.3}&17.0&12.5&33.8&65.5&61.6&\bf{52.0}&9.3&\bf{24.9}&54.1&\bf{49.1}&\bf{38.1}\\
\bottomrule[1.5pt]
\end{tabular}}
\end{table*}

\begin{table}[h]
\centering
\caption{Results on PASCAL VOC in adaptation from PASCAL VOC to Clipart Dataset. Average precision (\%) is evaluated on PASCAL. Our method does not degrade the performance on the source whereas BDC-Faster and DC-Faster degrade it.}
\label{tbl:ap_pascal}
\scalebox{0.99}{
  \tabcolsep=2.5pt
  \centering
\begin{tabular}{c|cccc|c}
\toprule[1.5pt]
%  & \multicolumn{20}{c}{AP for each class} & \\
%                \cmidrule(r){2-21}  
Method  & G & I & CTX & L & MAP  \\\hline
Source Only &&&&  &77.5\\
BDC-Faster &$\checkmark$&&&& 73.6 \\
DA-Faster &$\checkmark$&$\checkmark$&& &66.4\\\hline
%DA-Faster&$\checkmark$&$\checkmark$&&&&&&&&&&\hline
%DA-Faster &$\checkmark$&&&\\\hline
\multirow{3}{*}{Proposed} &$\checkmark$&&&& \bf{78.0} \\
 &$\checkmark$&&$\checkmark$&&77.6\\
 &$\checkmark$&&$\checkmark$&$\checkmark$&77.0\\
\bottomrule[1.5pt]
\end{tabular}
}
\end{table}
\vspace{-2mm}
\subsection{Context Vector based Regularization}\label{sec:context}
\vspace{-2mm}
%\ks{why is this called 'context'? what is the motivation here? what is being 'regularized' exactly, are you penalizing the weights to be smaller? also, are the domain classifier features used at test time to detect objects?}
We further propose a regularization technique to improve the performance of our model. 
As discussed above, regularizing the domain classifier with the segmentation loss was effective for stabilizing the adversarial training in domain adaptive segmentation~\cite{learnfromsynth}. The authors designed a domain classifier that outputs both the domain label and a semantic segmentation map. 
Motivated by this approach, we propose to stabilize the training of the domain classifier by the detection loss computed on source examples. We extract vectors $v_1$ and $v_2$ from the middle layers of the two domain classifiers respectively. These vectors should contain information about whole input image, which we call ``context''.
Then, we concatenate the vectors with all region-wise features as shown in Fig. \ref{fig:architecture} and train the domain classifiers to minimize the detection loss on source examples as well as minimize domain classification loss. During the test phase, the vectors are forwarded to obtain outputs.%To effectively regularize the domain classifier, we did not back-propagate the gradients of the loss to Faster RCNN module. If we do that, the module may help to extract discriminative features, and regularization will be unsuccessful.
\vspace{-2mm}
\subsection{Overall Objective}
\vspace{-2mm}
We denote the objective of detection modules as $ \mathcal{L}_{det}$, which contains the loss for region proposal networks and final classification and localization error. %The domain classifier takes features before region proposal networks. The domain classifier outputs 2-dimensional vector per one example and trained to predict domain label. 
The adversarial loss $ \mathcal{L}_{adv} (F,D)$ is summarized as, 
\begin{align}
 \mathcal{L}_{adv} (F,D) =  \mathcal{L}_{loc} (F_{1},D_{l}) +  \mathcal{L}_{global} (F,D_{g}) 
 \end{align}
Combined with the loss of detection on source examples, the overall objective is, 
\begin{align}\label{eq:final}
  \max_{D}\min_{F,R}  \mathcal{L}_{cls} (F,R) - \lambda  \mathcal{L}_{adv} (F,D) 
\end{align}
where $\lambda$ controls the trade-off between detection loss and adversarial training loss. The sign of gradients is flipped by a gradient reversal layer proposed by \cite{ganin2014unsupervised}. Each mini-batch has one labeled source and one unlabeled target example.% and parameters are udpated by following Eq. \ref{eq:final}.%In experiments, we will show the behavior of the model with the change in the value of $\gamma$. %Without specific notation, we set $ \lambda = 1.0$. 

\section{Experiments}
We evaluate our approach on four domain shifts--PASCAL \cite{everingham2010pascal} to Clipart \cite{inoue2018cross}, PASCAL to Watercolor \cite{inoue2018cross},  Cityscapes~\cite{cordts2016cityscapes} to FoggyCityscapes~\cite{sakaridis2018semantic}, and GTA~\cite{johnson2016driving} to Cityscapes--to demonstrate that it is effective for adaptation between both dissimilar and similar domains. Additionally, we provide experiments to verify our claim that complete feature matching can degrade the performance of the model in the target domain. 
%We did experiments on four adaptation scenarios. First, we will show the detail of implementations. Then, we will explain the details of experiments and results.

\noindent
\textbf{Implementation Details.}
In all experiments, we set the shorter side of the image to 600 following the implementation of Faster RCNN \cite{faster} with ROI-alignment \cite{maskrcnn}. We first trained the networks with learning rate $0.001$ for 50K iterations, then with learning rate $0.0001$ for 20K more iterations and reported the final performance. All models are trained with this scheduling and we reported the performance trained after 70K iterations. Without specific notation, we set $\lambda$ as 1.0 and $\gamma$ as 5.0. We implemented all methods with Pytorch \cite{pytorch}. Please see our supplemental material for the detail of  the network architecture.

We compared our method with three baselines: FRCNN model,  FRCNN with a baseline domain classifier, and domain adaptive FRCNN (DA-Faster)~\cite{dafaster}.  
FRCNN model was trained  only on source examples without any adaptation. 
The FRCNN with a baseline domain classifier has exactly the same architecture as our proposed weak-global alignment model, but its domain classifier is trained with cross-entropy loss in Eq. \ref{eq:fl_t} and \ref{eq:fl_s}. The model does not have a local-level domain classifier. By comparing with this model, we can directly observe the effectiveness of our proposed weak alignment approach. Hereafter, we call the baseline {\sl BDC-Faster}. 
\textit{DA-Faster}~\cite{dafaster} employs two domain classifiers, an image-level one for high-level features and an instance-level one for features cropped by the region proposal network. Both domain classifiers are trained by cross-entropy loss. In addition, it utilizes a technique called consensus regularization, which makes the outputs of two domain classifiers similar. Since we did not observe any benefit of the technique, we report the results without it.
Since we implemented the method ourselve, the results reported in the original paper and in our paper are different. We denote their reported performance as \textit{DA-Faster*}.

\def\subfigcapskip{5pt}
\begin{figure*}
  \centering
     \begin{tabular}{cccc}
 
 \subfigure[Proposed (MAP: 36.4)]{\includegraphics[width=0.3\textwidth]{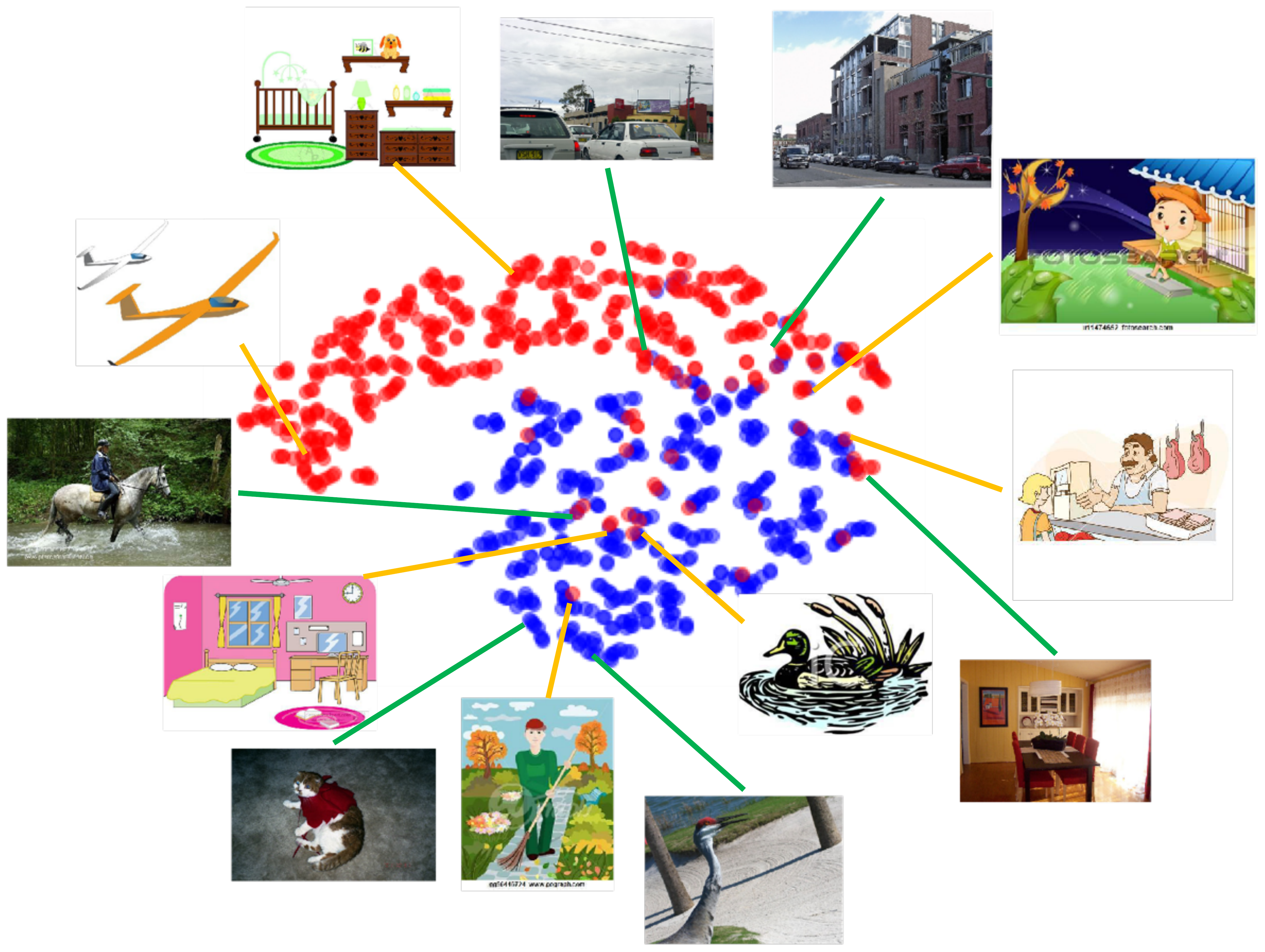}\label{fig:proposed_pascal}} 
     \subfigure[Baseline DC (MAP: 25.6)]{ \includegraphics[width=0.25\textwidth]{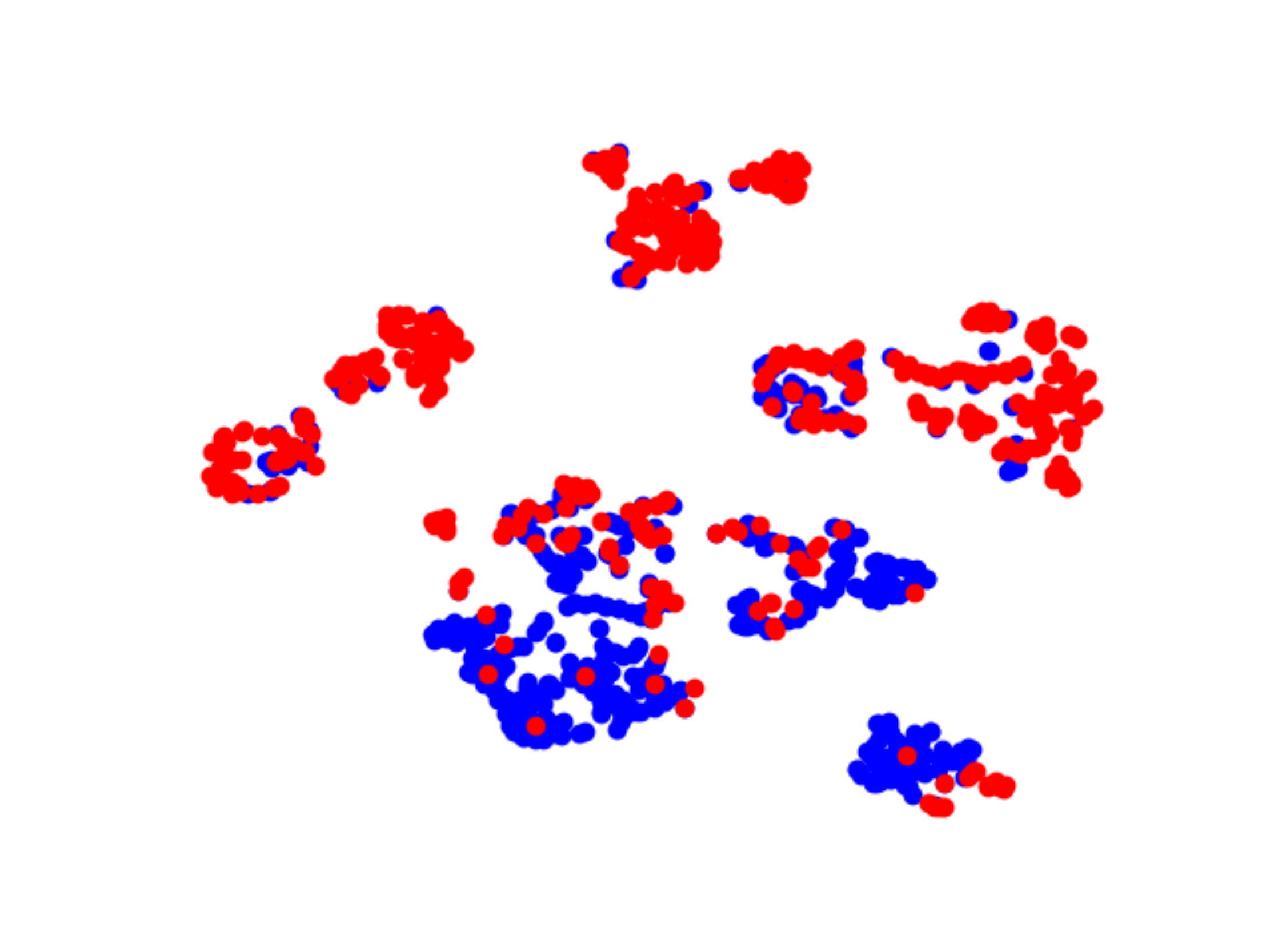}\label{fig:baseline_pascal}} 

  \subfigure[Proposed (MAP: 29.1)]{\includegraphics[width=0.22\textwidth]{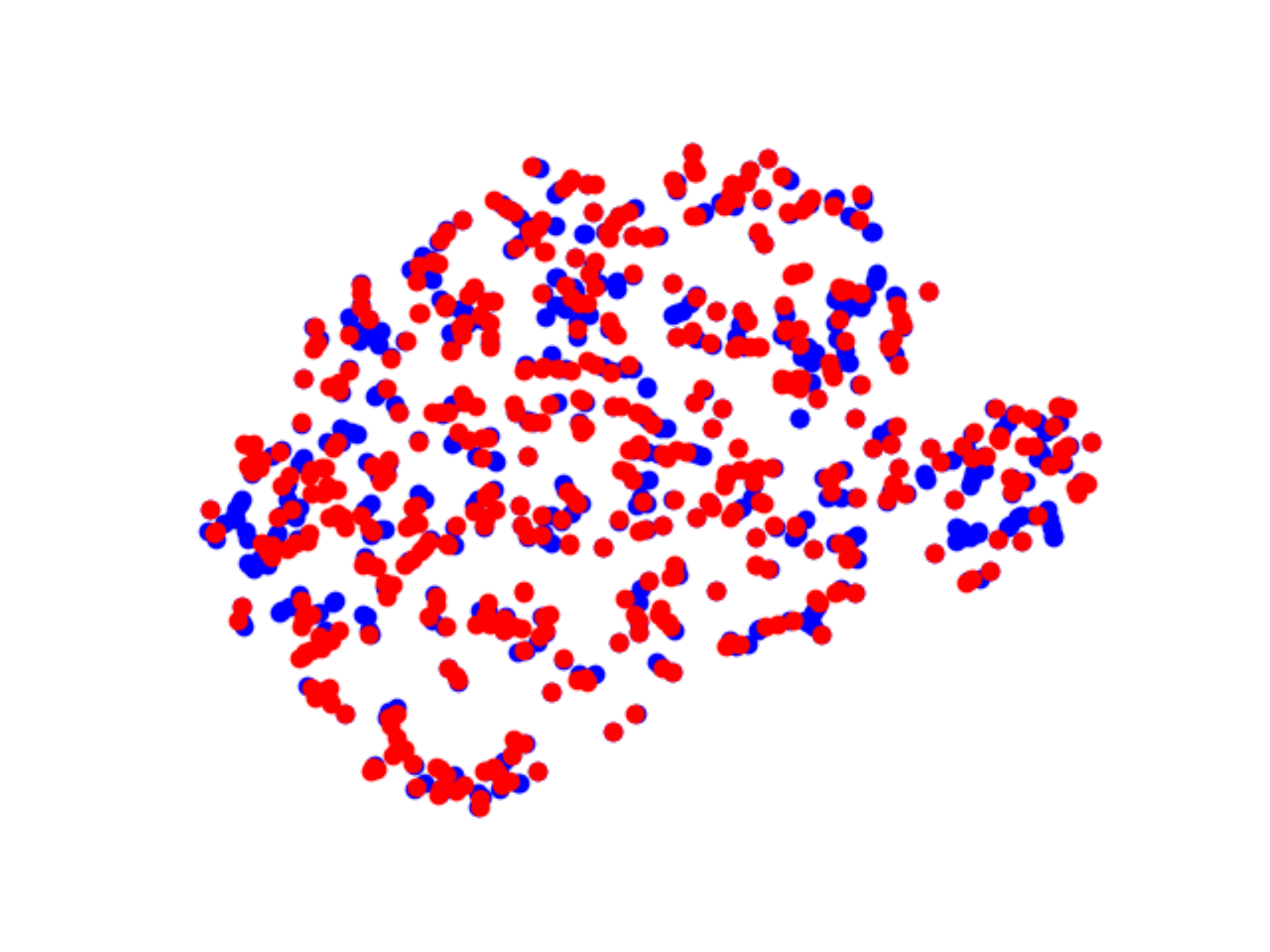}\label{fig:focal_city}}
   \subfigure[baseline DC (MAP: 27.6)]{\includegraphics[width=0.22\textwidth]{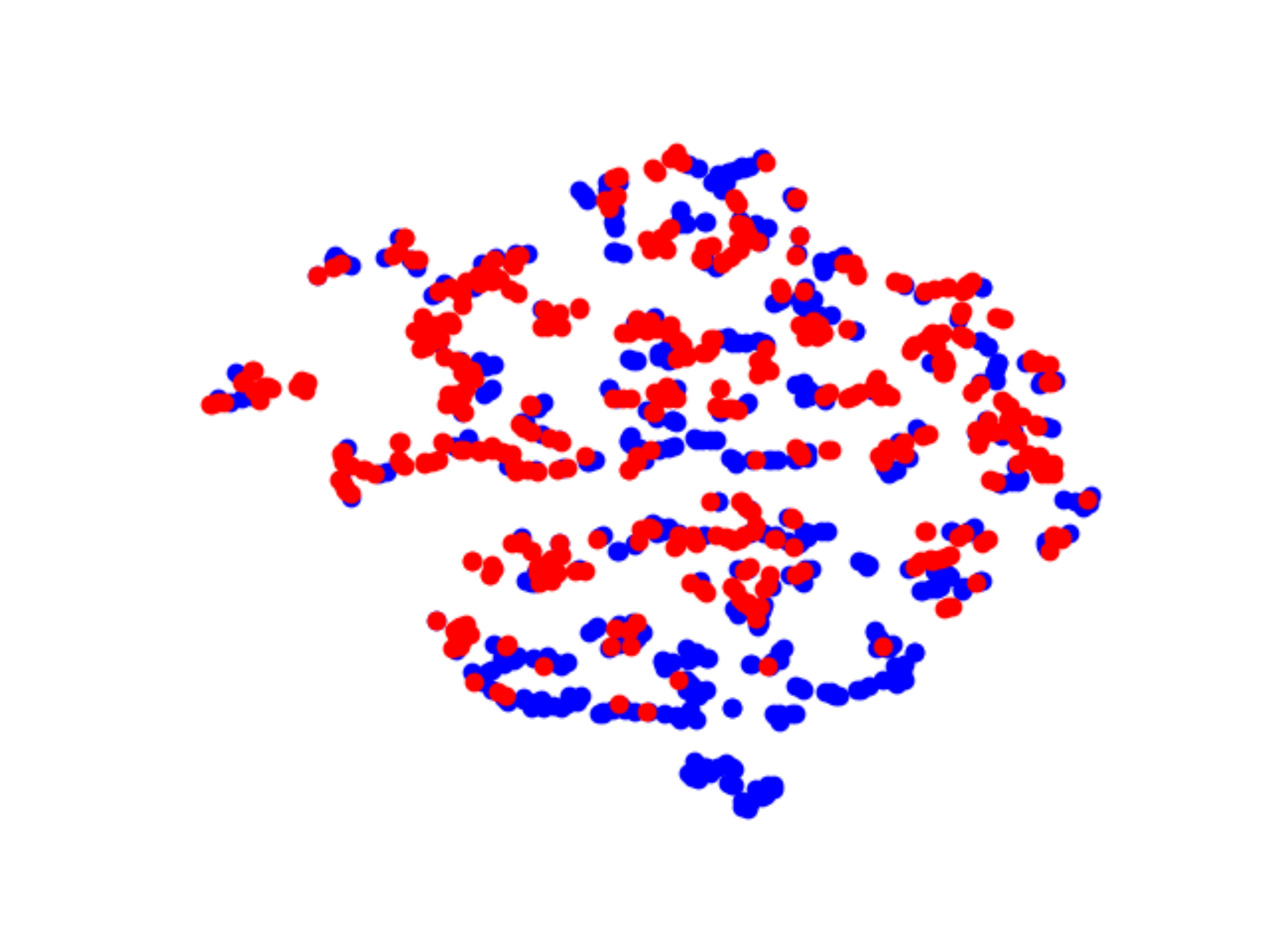}\label{fig:baseline_city}}%\end{minipage}
          \end{tabular}
           \caption{Visualization of features obtained by two different models. Blue: source examples, Red: target examples. Fig. \subref{fig:proposed_pascal} and \subref{fig:baseline_pascal} are the results of adaptation between dissimilar domains (from pascal to clipart). For Fig. \subref{fig:proposed_pascal}, images with green lines are from PASCAL VOC (source). Images with orange lines are from Clipart (target). Our method does not match feature distributions strictly whereas the baseline method matches. However, our method outperformed the baseline with a large margin, which demonstrates the effectiveness of global-weak alignment.
           Fig. \subref{fig:baseline_city} and \subref{fig:focal_city} are adaptation between similar domains (from Cityscape to FoggyCityscape). When the domains are very similar, the baseline method works well though our method performs better.}
         \label{fig:feautures_foggy}
\end{figure*}

\subsection{Adaptation between dissimilar domains}
We first show experiments on dissimilar domains, specifically, adaptation from real images to artistic images. 
We utilized the Pascal VOC Dataset as the real source domain \cite{everingham2010pascal}. This dataset contains 20 classes of images and their bounding box annotations. Following a prevalent evaluation protocol, we employed PASCAL VOC 2007 and 2012 training and validation splits for training, resulting in about 15k images. 
The target domain consists of either the Clipart or the Watercolor datasets~\cite{inoue2018cross}. Clipart contains comical images whereas Watercolor has artistic images.
Clipart contains 1K images in total, which have the same 20 categories as PASCAL VOC. 
All images were used for both training (without labels) and testing.
Watercolor contains 6 categories in common with PASCAL and 2K images in total. 1K training images were utilized during training and our model is evaluated on 1K test images. 
In this experiment, we used the ResNet101~\cite{he2016deep} pre-trained on~\cite{imagenet} as a backbone network. For other details see our supplemental material.  

\noindent
\textbf{Results on Clipart.}
As shown in Table \ref{tbl:ap_clipart}, our proposed method outperformed all baselines. Just by replacing the domain classifier's objective with the focal loss, MAP improved by 10.8$\%$ (25.6 to 36.4). In addition, the context vector based regularization and local alignment (C, L in Table), further improved MAP. The performance on the source domain, PASCAL VOC, is shown in Table \ref{tbl:ap_pascal}. Compared with the performance of the source only model, BDC-Faster and DA-Fastster significantly decrease its performance. This fact indicates that strictly aligning feature distributions between different domains can disturb the training for object detection while our method does not degrade the performance on the source domain. 

We further visualized the features obtained by two models, our proposed global-level adaptation model and BDC-Faster in Fig. \ref{fig:proposed_pascal} and \ref{fig:baseline_pascal}. The target features obtained by a baseline domain classifier are matched compactly with the source domain (Fig. \ref{fig:baseline_pascal}). On the other hand, with our proposed method (Fig. \ref{fig:proposed_pascal}), some features are aligned with the source features, but most of them are separated from source features. Source images usually focus on one or two objects whereas target images usually contain multiple images. Some target images focusing on single object are likely to be aligned with source as shown in the figure. 
Many existing methods for image classification aimed to match the feature distributions closely. However, this visualization implies that such distribution matching does not always help domain adaptive object detection.

\noindent
\textbf{Results on Watercolor.}
According to Table \ref{tbl:ap_water}, our method outperformed the baseline methods. There was a large improvement on this domain. The improvement by the local alignment is especially large, about 3$\%$, because the target images have a characteristic ``painting'' style. Therefore, the reducing the domain-gap based on local-level features  improves the performance. %Different from the results on clipart dataset, a baseline domain classifier based method performed a little better than the source only model. %The possible reason is that images of Watercolor dataset descr%are more similar to real images than those of Clipart dataset are. 

\begin{table}[]
\centering
 \caption{AP on adpatation from PASCAL VOC to WaterColor ($\%$). The definition of G, I, CTX, L is following Table \ref{tbl:ap_clipart}.} 
 \label{tbl:ap_water}
 \scalebox{0.85}{
 \tabcolsep=1.5pt
\begin{tabular}{l|cccc|ccccccc}
\toprule[1.5pt]
	&\multicolumn{4}{c|}{}  & \multicolumn{7}{c}{AP on a target domain}\\%\cmidrule(r){6-19}
Method   & G & I & CTX & L & bike & bird & car  & cat  & dog  & prsn & MAP  \\\hline
Source Only &&&& &68.8 & 46.8 & 37.2 & 32.7 & 21.3 & 60.7   & 44.6 \\
BDC-Faster&$\checkmark$&& &&68.6 & 48.3 & 47.2 & 26.5 & 21.7 & 60.5   & 45.5\\
DA-Faster&$\checkmark$&$\checkmark$&& &75.2 & 40.6 & \bf{48.0} & 31.5 & 20.6 & 60.0   & 46.0\\\hline
\multirow{3}{*}{Proposed} &$\checkmark$&&& &66.4 & 53.7 & 43.8 & \bf{37.9} & 31.9 & 65.3   & 49.8 \\
&$\checkmark$&& $\checkmark$&& 71.3 & 52.0 & 46.6 & 36.2 & 29.2 & \bf{67.3}   & 50.4\\
&$\checkmark$&& $\checkmark$& $\checkmark$& \bf{82.3}	& \bf{55.9}&46.5&32.7&\bf{35.5}&66.7&\bf{53.3}\\
\bottomrule[1.5pt]
\end{tabular}}
\end{table}

\begin{table}[]
 \caption{AP on adaptation from Cityscape to FoggyCityscape ($\%$). The performance of our method is very near to oracle, which is trained on labeled target images.} 
 \label{tbl:ap_foggy}
 \centering
 \scalebox{0.8}{
 \tabcolsep=1.5pt
\begin{tabular}{l|cccc|ccccccccc}
\toprule[1.5pt]
%&  && 	& \multicolumn{8}{c}{AP for each class} & \\ \cmidrule(r){2-12}
	&\multicolumn{4}{c|}{}  & \multicolumn{9}{c}{AP on a target domain} \\
Method& G & I & CTX & L &bus  & bcycle & car  & bike & prsn & rider & train & truck & MAP \\\hline
\scriptsize{Faster RCNN}& &&&&  22.3 & 26.5   & 34.3 & 15.3   & 24.1   & 33.1  & 3.0   & 4.1   & 20.3 \\
\scriptsize{BDC-Faster}&$\checkmark$  &&& &29.2&28.9&42.4&22.6&26.4&37.2&12.3&21.2&27.5\\
\scriptsize{DA-Faster}&$\checkmark$  &$\checkmark$&&& 33.1&23.3&25.5&15.6&23.4&29.0&10.9&19.6&22.5 \\
\scriptsize{DA-Faster*}&$\checkmark$  &$\checkmark$&&& 25.0&31.0&40.5&22.1&\bf{35.3}&20.2&20.0&\bf{27.1}&27.6 \\\hline

\multirow{4}{*}{Proposed}&$\checkmark$  &&&&   33.5 & 33.3   & 42.7 & 22.2   & 27.1   & 40.3  & 11.6  & 22.3  & 29.1 \\
&&&&$\checkmark$&34.3&32.2&36.2&23.7&27.5&39.3&5.4&24.4&27.9\\
&$\checkmark$  &&$\checkmark$ && \bf{ 38.0} & 31.2   & 41.8 & 20.7   & 26.6   & 37.6  & 19.7  & 20.5  & 29.5\\
&$\checkmark$  &&$\checkmark$ &$\checkmark$& 36.2&\bf{35.3}&\bf{43.5}&\bf{30.0}&29.9&\bf{42.3}&\bf{32.6}&24.5&\bf{34.3}\\\hline
Oracle &&&&&50.0&36.2&49.7&34.7&33.2&45.9&37.4&35.6&40.3\\
\bottomrule[1.5pt]
\end{tabular}}
\end{table}
\subsection{Adaptation between similar domains}
In this experiment, we aim to analyze our method by evaluating the adaptation between very similar domains. 
We used Cityscape \cite{cordts2016cityscapes} as the source domain. %The dataset is designed to develop the method for scene understanding for driving scenarios. 
The images in the dataset are captured by a car-mounted video camera.
As the target domain, we used FoggyCityscape datasets \cite{sakaridis2018semantic}. The images are rendered from Cityscape using depth information and it simulates the change of weather condition. The important difference from other adaptation scenario is that source and target images are originally the same one. Target images are generated from source images by adding fog noise. In such adaptation scenario, strictly aligning feature distributions should be effective because there exists a correct matching between source and target images. 
Both dataset have 2, 975 images in the training set, and 500 images in the validation set. We utilized the training set during training and evaluated on the validation set. 
Since Cityscapes dataset does not have bounding-box annotation, we take the tightest rectangles of its instance masks as groundtruth bounding boxes. We used the VGG16 model \cite{simonyan2014very} as a backbone network following~\cite{dafaster}.

As shown in Table~\ref{tbl:ap_foggy}, our proposed method performed much better than the baseline methods. MAP of a model with only strong local alignment was 27.9. Combining strong local and weak global alignment boosted MAP to 34.3. 
The domain-shift is caused by fog noise, a local-level shift. Hence, strong local alignment largely contributed to the improvement. 
In this adaptation scenario, the method with a baseline domain classifier performs better than the source only model. This is because the target images have exactly the same layout and number/combination of objects. Thus, strong alignment between different domains was effective. The visualized features in Fig.~\ref{fig:feautures_foggy} show completely different characteristics from the experiments on PASCAL to Clipart dataset. The features are matched in both methods. The results indicate that our proposed method performs both when two domains are dissimilar and similar.

\begin{figure*}
    \centering
    \includegraphics[width=0.95\textwidth]{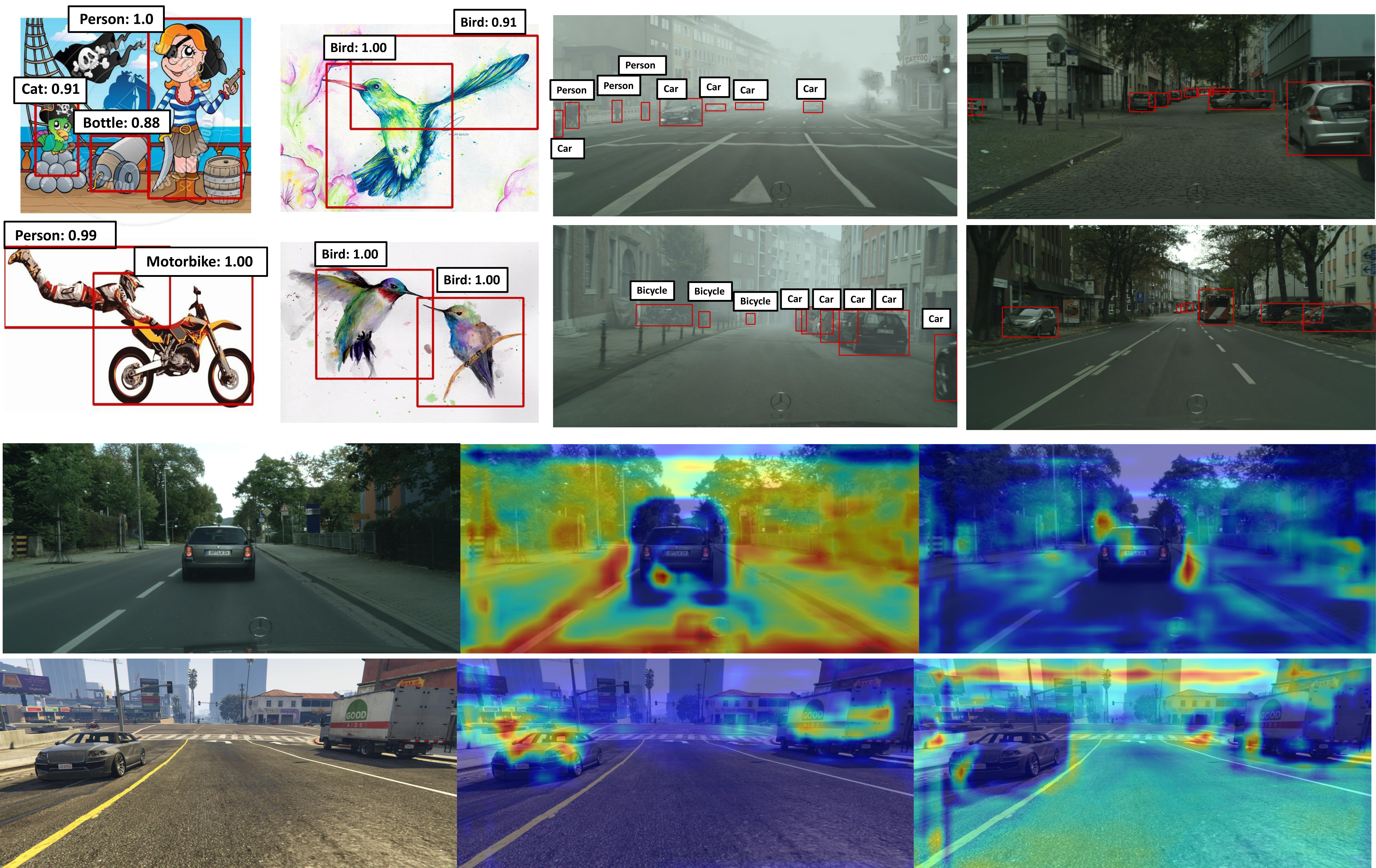}
    \caption{Upper: Examples of detection results on the target domain. From left to right column, Clipart, Watercolor, FoggyCityscape and Cityscape dataset. Bottom: Visualization of domain evidence using Grad-Cam. The evidence is obtained by the global-domain classifier. The pictures show results on target (Top) and source images (Bottom). From left to right, input images, images of evidence for the target, evidence of the source domain. The feature extractor seems to focus on deceiving the domain classifier in regions with cars.}
    \label{fig:examples_result}
\end{figure*}

\begin{table}[t!]
 \caption{Results on adpatation from Sim10k to Cityscape Dataset ($\%$). Average precision is evaluated on target images. FL ($\gamma=3$)* indicates the experiments in which shorter side of image is scaled to 1000 during training and testing. {\textit P} indicates pixel-level alignment, whether we used images generated by cyclegan during training. $\dagger$ indicates the performance when the context vector is zero-padded and not used for the output.}
 \label{tbl:ap_sim10k}
 \centering
 \scalebox{0.85}{
\begin{tabular}{l|ccccc|c}
\toprule[1.5pt]
%&  && 	& \multicolumn{8}{c}{AP for each class} & \\ \cmidrule(r){2-12}
Method& G & I & CTX & L &P & AP on Car  \\\hline
Faster RCNN& &&&&&  34.6  \\
BDC-Faster&$\checkmark$  &&&&&  31.8 \\
DA-Faster&$\checkmark$  &$\checkmark$&&&& 34.2\\
DA-Faster*&$\checkmark$  &$\checkmark$&&&& 38.9\\
Weak Align&$\checkmark$  &$\checkmark$&&&& 35.8 \\\hline
\multirow{5}{*}{Proposed (FL)}&$\checkmark$  &&&&&   36.4 \\
&$\checkmark$  &&$\checkmark$ &&&  38.2  (38.3$\dagger$)	 \\
&$\checkmark$  &&$\checkmark$&$\checkmark$ &&  40.1 \\
 &$\checkmark$  &&$\checkmark$&&$\checkmark$&  41.5 \\
 &$\checkmark$  &&$\checkmark$&$\checkmark$&$\checkmark$&  40.7 \\\hline
 \multicolumn{7}{c}{Proposed Method with different parameters} \\\hline
EFL &$\checkmark$  &&$\checkmark$ &&&  38.7 \\%\hline
FL ($\gamma=3$) &$\checkmark$  &&$\checkmark$ &&&  42.3 \\
FL ($\gamma=3$)*&$\checkmark$  &&$\checkmark$ &&$\checkmark$&  47.7 \\\hline
Oracle &&&& &&  53.1 \\
\bottomrule[1.5pt]
\end{tabular}}
\end{table}

\subsection{Adaptation from synthetic to real images}
We evaluate the performance of our model in an adaptation from synthetic images to real images. As the synthetic domain, we used Sim10k \cite{johnson2016driving}. The dataset contains images of the synthetic driving scene, 10,000 training images which are collected from the computer game Grand Theft Auto (GTA). We employed the same architecture as used in the previous section. Following the protocol of \cite{dafaster}, we evaluated detection performance on \textit{car}. As a real domain, we used Cityscape. All training images are used during training for both domains. Average precision was evaluated on the validation split of the Cityscape. We set the value of $\lambda = 0.1$ following \cite{dafaster} in Eq. \ref{eq:final}. We show the performance when varying the value of $\lambda$ in our supplemental material. The two domains have similar layout in that both domains are driving scene images. However, the color and lighting are clearly different. In this respect, the two domains are more different than Cityscape and Foggycityscape are.
We extensively evaluated our method by ablating some components.  Moreover, we show the results using instance-level adaptation as proposed in \cite{dafaster}. We also show the comparison and results of combination with a model trained with images translated by CycleGAN \cite{zhu2017unpaired}. We trained CycleGAN to translate different domains' images, then utilized the translated source images for training. Whether we employed the translated images is denoted by the colum of P in Table \ref{tbl:ap_sim10k}. The details are shown in supplemental material. 
In addition, we demonstrate that our idea of weak alignment can be achieved with a loss function other than focal loss. In Eq. \ref{eq:decrease}, we set  $f(p_t) = e^{-\eta p_t}$, which is a decreasing function with the value of $p_t$. We call the loss function exponential focal loss (EFL). We set $\eta=5.0$.

The results are summarized in Table \ref{tbl:ap_sim10k}. Our method constantly performed better than the baseline models. Comparing the results of BDC-Faster (31.8) and our method with only global-level alignment (36.4), the weak feature distribution alignment outperformed the strict alignment. Setting the value of $\gamma=3.0$ in Focal Loss significantly improved the performance. In addition, with regard to a model trained with EFL, we could observe the improvement over the baseline models. The results demonstrate that our idea of weak global alignment is effective and can be achieved by functions other than Focal Loss. 

Context vector based regularization and local-level alignment further improved the performance. The performance did not degrade when we did not use the context vector in test phase as seen in the table. This implies that the network does not use the vector for the prediction whereas the performance improved compared to the model without the regularization. Therefore, the context vector seems to contribute to the regularization of the domain classifier.

We could not see a positive effect of instance-level adaptation (Weak Align in Table \ref{tbl:ap_sim10k}). Instance-level alignment utilizes the cropped features by region proposal networks, but the proposals may not localize objects in the target domain well, so it can hurt the performance of the model.

\subsection{Analysis}
\noindent
\textbf{Examples of detection results.}
We show the examples of detection results in Fig.~\ref{fig:examples_result}. Even when the style of the images is different between the source and target, our model localizes objects correctly in these cases. As seen in Clipart's example, when the appearance of the objects is largely different, the detection results are not successful. Also, as seen in case of Watercolor, the detector tends to output multiple predictions to one object. In case of FoggyCityscape's examples, our model tends to assign one bounding box to multiple neighboring bicycles. 

\noindent
\textbf{Visualization of domain evidence.}
To analyze the behavior of the feature extractor and domain classifier, we visualize the evidence for the global-level domain classifier's prediction using Grad-cam \cite{grad-cam} in Fig.~\ref{fig:examples_result}. We use Grad-cam to show the evidence (heatmap) for why the domain classifier thinks the image comes from the source or the target, for the adaptation from Sim10k to Cityscapes. Please see our supplemental material for other examples. For the target images, the domain classifier does not look at cars as the evidence for the target. Similarly, for source images, it also does not look at cars as the evidence for the source. This indicates that the feature extractor seems to focus on cars to deceive the domain classifier, which means that the feature extractor learns to partially align global-image features, specifically around cars. 

\section{Conclusion}
\vspace{-2mm}
In this work, we propose a novel approach for detector adaptation based on strong local alignment and weak global alignment for unsupervised adaptation of object detectors. Our key contribution is the weak alignment model, which focuses the adversarial alignment loss on images that are globally similar and puts less emphasis on aligning images that are globally dissimilar. 
Additionally, we design the strong domain alignment model to only look at local receptive fields of the feature map. Our method outperformed other existing methods with a large-margin in several datasets. Through extensive experiments, we verified the effectiveness of weak global and strong local alignment. 

\section{Acknowledgements}
This work was supported by Honda, DARPA and NSF Award No. 1535797 and partially supported by JST CREST Grant Number JPMJCR1403, Japan.

{\small
\bibliographystyle{ieee}
\bibliography{egbib}
}

\onecolumn
\setcounter{section}{0}
\section*{Supplemental Material}
\setcounter{section}{0}
\section{Network Architecture}
\vspace{-3mm}
We used the same architecture of the domain classifier for Faster RCNN with ResNet101 and VGG16. As the number of the channels in input features is different, we changed the channel size of network according to the backbone network. 

In case of VGG16 model, the feature in $conv3\_3$ layer is fed into the local domain classifier. 
The feature in the last $res2c$ layer is fed into the local domain classifier with regard to ResNet101 model. The name of the layer is cited by the Caffe~\cite{caffe} prototxt.
\vspace{-3mm}
\begin{table}[h!t]
\caption{The architecture of the domain classifiers.}
\begin{minipage}[t]{.5\textwidth}
%\footnotesize
\begin{center}
\begin{tabular}{|c|}
\hline
\multicolumn{1}{|c|}{Global Domain Classifier} \\
\hline
%Input: Features  \\\hline
%\multicolumn{1}{|c|}{Generator} \\\hline
	Conv $3 \times 3 \times 512$, stride 2, pad 1\\
	Batch Normalization, ReLU, Dropout\\
    Conv $3 \times 3 \times 128$, stride 2, pad 1\\  
    Batch Normalization, ReLU, Dropout\\
    Conv $3 \times 3 \times 128$, stride 2, pad 1\\  
    Batch Normalization, ReLU, Dropout\\
    Average Pooling\\
	Fully connected $128 \times 2$\\
	Softmax\\\hline
\end{tabular}
\end{center}
\end{minipage}
\begin{minipage}[t]{.5\textwidth}
%\footnotesize
\begin{center}
\begin{tabular}{|c|}
\hline
\multicolumn{1}{|c|}{Local Domain Classifier} \\
\hline
%Input: Features  \\\hline
%\multicolumn{1}{|c|}{Generator} \\\hline
	Conv $1 \times 1 \times 256$, stride 1, pad 0\\
	ReLU\\
    Conv $1 \times 1 \times 128$, stride 1, pad 0\\  
    ReLU\\
    Conv $1 \times 1 \times 1$, stride 1, pad 0\\  
    Sigmoid\\\hline
\end{tabular}
\end{center}
\end{minipage}
\end{table}

\noindent
\textbf{Global Domain Classifier}\par
The global domain classifier has three layered convolution layers, global average pooling and one Linear layer. The kernel size of the convolution layers is set as three. Batch Normalization, ReLU, and dropout layers are attached after each convolution layer. The output of the global domain classifier is activated by softmax function. Context vector is extracted after the average pooling layer. Therefore, the vector has 128 dimensions.

\noindent
\textbf{Local Domain Classifier}\par
The local domain classifier has three layered convolution layers. The kernel size of the convolution layers is set as one. The output of the local domain classifier is activated by sigmoid function. Context vector is extracted before the last convolution layer. Therefore, the vector has 128 dimensions.

\section{Pixel-level Adaptation}
The results on a model trained with images generated by CycleGAN were provided in our paper. We describe the details of how we trained the model. 

\noindent
\textbf{Training of CycleGAN}\par
To train the CycleGAN, we used the Pytorch implementation {\scriptsize \url{https://github.com/junyanz/pytorch-CycleGAN-and-pix2pix}. }
We used all training images in both domains. We trained CycleGAN for 10 epochs and employed the source images translated into the target domain for training our Faster RCNN model.

\noindent
\textbf{Training of Faster RCNN}\par
We found that some of the translated images are not translated correctly. Cars are completely hidden by large noise. In order to suppress the effect of such corrupted images, we trained our model using source, translated source and target images. Translated images are utilized just for training detection modules and not utilized for domain classification. Namely, we have three images in each mini-batch, source, translated source and target image. Source and target images are used as we mentioned in our main paper. The translated source one is used to calculate and back-propagate the detection loss.

\section{Additional Results}
%We provide additional results that are not shown in our main paper due to the limit of space. 

\noindent
\textbf{Results on source domain}\par
Table \ref{tbl:ap_pascal2}, \ref{tbl:water_all2}, and \ref{tbl:ap_foggy2} show the results on source domain in three adaptation scenarios. In all scenarios, our method does not significantly degrade the detection performance on the source domain.

\noindent
\textbf{Results on target domain}\par
Table \ref{tbl:clipart_all2}, \ref{tbl:water_all2} and \ref{tbl:ap_sim10k2} provide results including local-level only adaptation and pixel-level adaptation. We did not show the pixel-level adaptation results on Clipart and Watercolor dataset in our main paper. With regard to Cityscape, we show more ablations in this table.
In adaptation for clipart dataset, training with the images generated by CycleGAN does not improve the performance. The possible reasons are that the style of the target images is largely different and that target images have diverse styles of examples. Then, CycleGAN may not generate images suitable for adaptive detection. 
In the experiments on Watercolor, the performance greatly improved with the use of pixel-level adaptation. The model demonstrates almost oracle-level performance. Table \ref{tbl:clipart_vgg} denotes the results when using VGG network for the adaptation from PASCAL VOC to clipart. The performance largely improved with our method.

\noindent
\textbf{Parameter Sensitivity}\par
\begin{figure}[t!]
\centering
 \subfigure[Parameter sensitivity to $\lambda$]{\includegraphics[width=0.45\hsize]{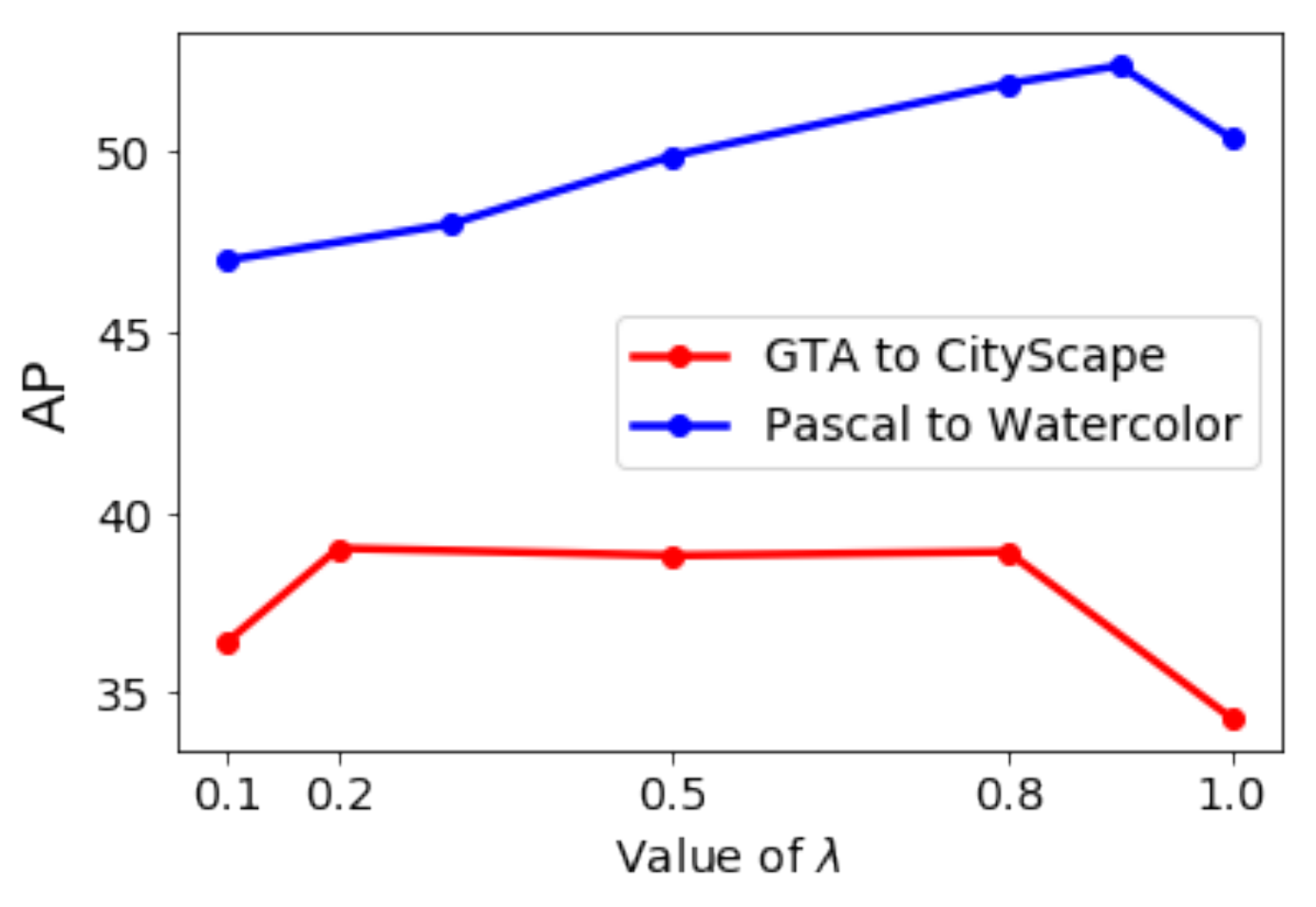} \label{fig:lambda_change}}
   \subfigure[Parameter sensitivity to $\gamma$]{\includegraphics[width=0.425\hsize]{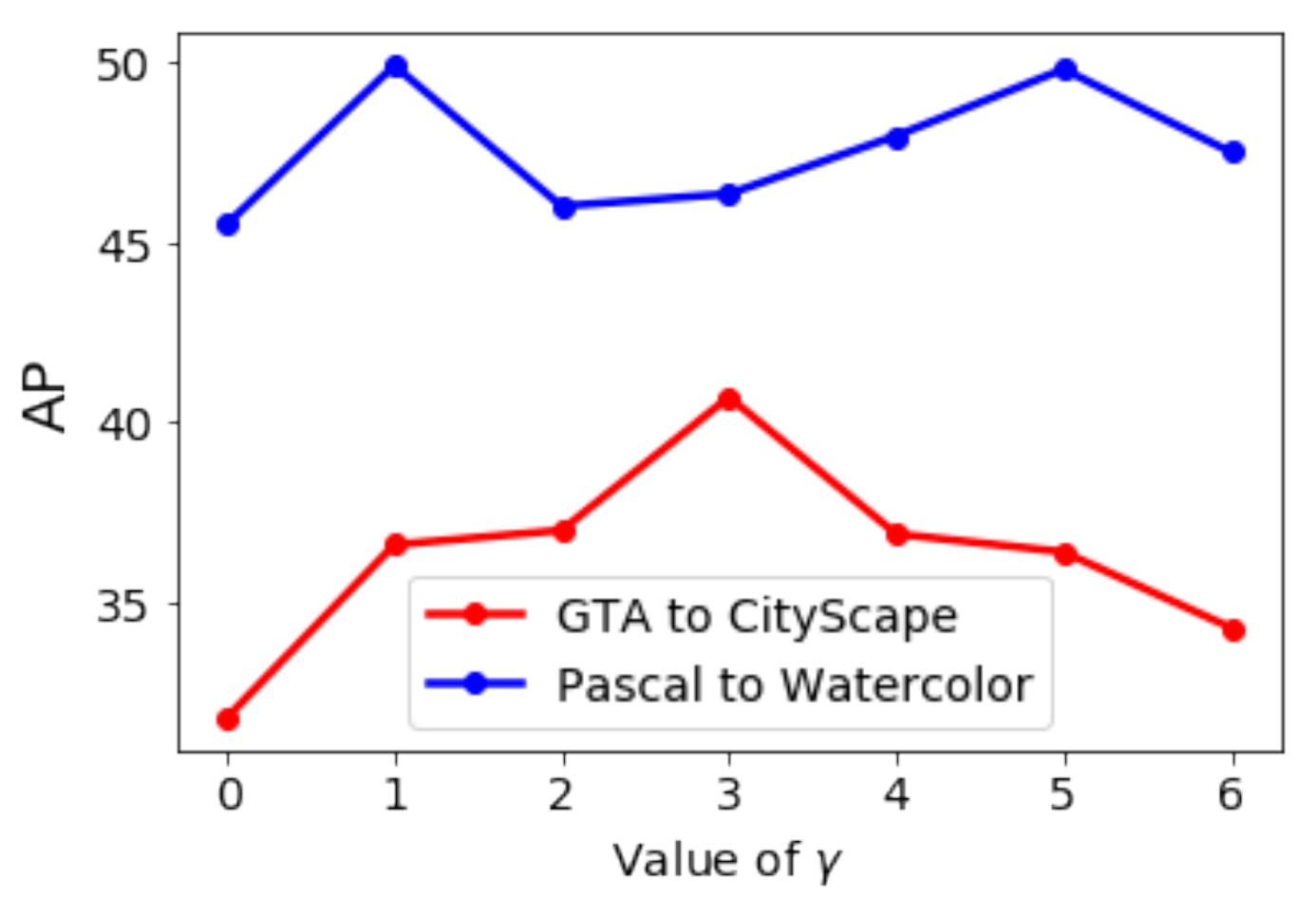}    \label{fig:gamma_change}
} 
      \caption{Parameter sensitivity to the value of $\lambda$ (Left) and $\gamma$ (Right) in adaptation from Sim10k to Cityscape and from Pascal to Watercolor.}
\end{figure}

Fig. \ref{fig:lambda_change} presents the sensitivity to the parameter $\lambda$ in Eq. 12 in our main paper. The parameter is the trade-off between training of detection and adversarial training. Our method performed better than the baseline domain classifier in all values ranging from $0$ to $1$.

Fig. \ref{fig:gamma_change} presents the sensitivity to the parameter $\gamma$ of Focal Loss. The result is obtained in adaptation from Sim10k to Cityscape. The parameter controls how strictly we align features between domains. As the parameter gets small, the domain classifier will look at all examples. As shown in the figure, the peak of the performance was around $3.0$, in which AP was 42.3. 

\noindent
\textbf{Domain Evidence}\par
\begin{figure}
    \centering
    \includegraphics[width=1.0\hsize]{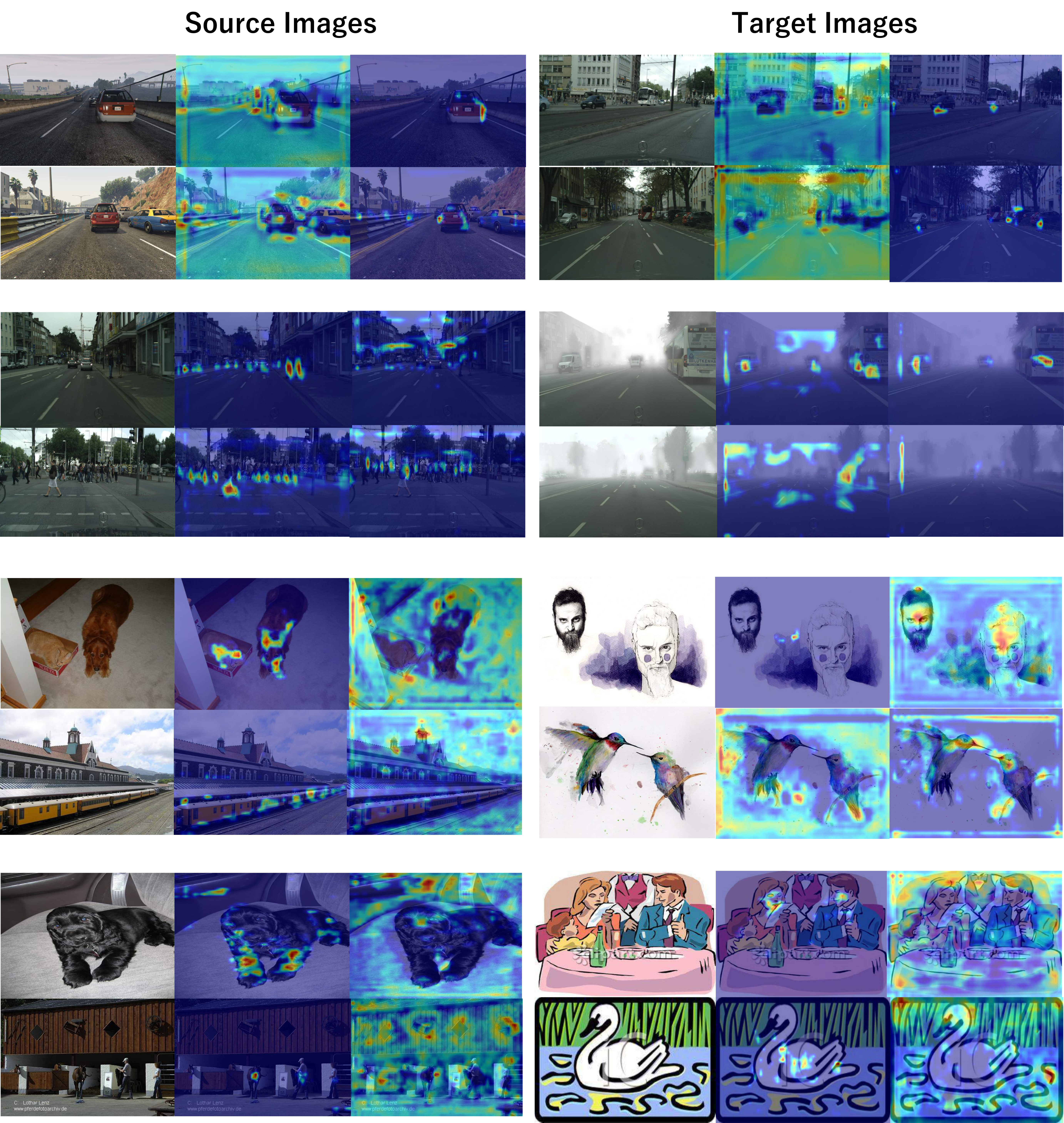}
    \caption{Visualization of domain evidence using Grad-Cam. The evidence is obtained by the global-domain classifier. The pictures show results on target and source images respectively. From left to right, input images, images of evidence for the target, evidence of the source domain. The behavior of the domain classifier seems to be different in the adaptation scenarios. However, the feature extractor tries to partially fool the domain classifier.}
    \label{fig:vis_gradcam2}
\end{figure}

Fig. \ref{fig:vis_gradcam2} is the additional domain evidence visualization. Although the behavior differs from dataset to dataset, the feature extractor tries to partially fool the domain classifier. %The evidence is extracted from either foreground or background.  

\begin{table*}[h]
\centering
\caption{Results on PASCAL VOC in adaptation from PASCAL VOC to Clipart Dataset. Average precision (\%) is evaluated on PASCAL. Our method does not degrade the performance on the source whereas BDC-Faster and DC-Faster degrade it.}

\label{tbl:ap_pascal2}

\scalebox{0.8}{
  \tabcolsep=1.5pt
  \centering
\begin{tabular}{c|cccc|ccccccccccccccccccccc}
\toprule[1.5pt]
%  & \multicolumn{20}{c}{AP for each class} & \\
%                \cmidrule(r){2-21}  
Method  & G & I & C & L & aero & bcycle & bird & boat & bottle & bus  & car  & cat  & chair & cow  & table & dog  & hrs & bike & prsn & plnt & sheep & sofa & train & tv & MAP  \\\hline
Faster RCNN &&&&  &77.7&80.3&82.5& \bf{79.0}& \bf{68.0}&88.1&85.7&87.0&53.1& \bf{87.3}&	58.2&88.3&85.0&87.9& \bf{80.5}& \bf{52.8}&75.9&69.4&86.3&77.8&77.5\\
BDC-Faster &$\checkmark$&&& &77.6      & 80.0    & 77.6 & 61.8 & 61.3   & 83.0 & \bf{86.1} & 86.5 & 50.9  & 79.8 & 59.8        & 84.7 & 82.2  & 79.5      & 78.1   & 45.3        & 73.4  & 70.1 & 80.8  & 73.8      & 73.6 \\
DA-Faster &$\checkmark$&$\checkmark$&& & 64.5&70.6&66.4&62.9&49.9&77.2&78.0&70.7&44.7&78.3&56.1&70.8&	75.6&85.2&74.5&37.2&65.4&60.2&72.3&66.9&66.4\\\hline
%DA-Faster&$\checkmark$&$\checkmark$&&&&&&&&&&\hline
%DA-Faster &$\checkmark$&&&\\\hline

\multirow{3}{*}{Proposed} &$\checkmark$&&&& \bf{79.5}      & 80.7    & 82.5 & 78.6 & 62.2   & 86.1 & 85.4 & 87.5 &  \bf{60.2}  & 79.0 & \bf{68.8}        & 88.6 & 86.2  &  \bf{88.7}      & 78.9   & 52.5        & 78.5  &  \bf{71.6} &  \bf{87.9}  & 76.9      &  \bf{78.0} \\
 &$\checkmark$&&$\checkmark$&&78.6&81.7& \bf{83.4}&74.7&62.9&86.9&85.4& \bf{90.6}&56.8&	85.7&57.9&89.0&87.4&87.7&78.1&52.3& \bf{79.3}&67.6&87.6& \bf{79.5}&77.6\\
 &$\checkmark$&&$\checkmark$&$\checkmark$&77.5& \bf{84.7}&81.1&71.1&63.8& \bf{88.5}&84.7&87.7&54.5&81.6&60.8& \bf{89.4}	& \bf{87.8}&88.2&78.2&49.9&78.9&70.2&82.0& \bf{79.5}&77.0\\

\bottomrule[1.5pt]
\end{tabular}
}
\end{table*}

\begin{table*}[h]
 \centering

\caption{Results on adaptation from PASCAL VOC to Clipart Dataset. Average precision (\%) is evaluated on target images. G, I, CTX, L, P indicate global alignment, instance-level alignment, context-vector based regularization, local-alignment and pixel-alignment respectively. Faster RCNN* indicates Faster RCNN trained on source images and source images translated by CycleGAN.}

\label{tbl:clipart_all2}

%\caption{Results on adaptation from PASCAL VOC to Clipart Dataset. Average precision (\%) is evaluated on target images. G, I, CTX, L, Pindicate global alignment, instance-level alignment, context-vector based regularization, local-alignment and pixel-alignment respectively. {Faster RCNN}^* indicates Faster RCNN trained by source images and translated source images by CycleGAN.}

%\label{tbl:clipart_all}
\scalebox{0.8}{
  \tabcolsep=1.5pt
  \centering
\begin{tabular}{c|ccccc|ccccccccccccccccccccc}
\toprule[1.5pt]
Method & G & I & CTX &L&P&aero & bcycle & bird & boat & bottle & bus  & car  & cat  & chair & cow  & table & dog  & hrs & bike & prsn & plnt & sheep & sofa & train & tv & MAP  \\\hline
Faster RCNN&&&&&&\bf{35.6}      & 52.5    & 24.3 & 23.0 & 20.0   & 43.9 & 32.8 & 10.7 & 30.6  & 11.7 & 13.8        & 6.0  & \bf{36.8}  & 45.9      & 48.7   & 41.9        &\bf{16.5}  & 7.3  & 22.9  & 32.0      & 27.8 \\
{Faster RCNN}*&&&&&$\checkmark$&26.4 & 52.7 & 28.3 & 24.1 & 28.5 & 49.7 & 30.2 & 13 & 35.3 & 26.5 & 15.8 & 7.6 & 26.1 & 68.1 & 47.2 & 42.5 & 5.9 & 23.2 & 41 & 42.6 & 31.7\\
BDC-Faster &$\checkmark$&&&&&20.2      & 46.4    & 20.4 & 19.3 & 18.7   & 41.3 & 26.5 & 6.4  & 33.2  & 11.7 &\bf{ 26.0}        & 1.7  & 36.6  & 41.5      & 37.7  & 44.5        & 10.6  & 20.4 & 33.3  & 15.5      & 25.6 \\
DA-Faster&$\checkmark$&$\checkmark$&&&&15.0&34.6&12.4&11.9&19.8&21.1&23.2&3.1&22.1&26.3&10.6&10.0&19.6&39.4&34.6&29.3&1.0&17.1&19.7&24.8&19.8
\\\hline

\multirow{5}{*}{Proposed} &$\checkmark$&&&&&30.5      & 48.5    & \bf{33.6} & 24.8 & 41.2   & 48.9 & 32.4 & 17.2 & 34.5  & 55.0 & 19.0        & 13.6 & 35.1  & 66.2      & \bf{63.0}   & 45.3        & 12.5  & 22.6 & 45.0  & 38.9      & 36.4 \\
 &&&&$\checkmark$&&19.8&50.7&25.4	&21.7&30.2&47.2	&27.1&	8.5&33.5&26.8&14.0&11.7&31.5&62.0&49.9&39.6&9.1&23.8&	39.5&38.4&30.5\\
 
 &$\checkmark$&&$\checkmark$&&&31.7      &\bf{55.2}    & 30.9 & 26.8 & \bf{43.4}   & 47.5 & \bf{40.0} & 7.9  &\bf{36.7}  & 50.0 & 14.3        & \bf{18.0} & 29.2  & \bf{68.1}      & 62.3   & 50.4        & 13.4  & 24.5 & \bf{54.2}  & 45.8      & 37.5\\ 
 &$\checkmark$&&$\checkmark$&$\checkmark$&&26.2&48.5&32.6&\bf{33.7}&38.5&\bf{54.3}&	37.1&\bf{18.6}&34.8&\bf{58.3}&17.0&12.5&33.8&65.5&61.6&\bf{52.0}&9.3&\bf{24.9}&54.1&\bf{49.1}&\bf{38.1}\\
  &$\checkmark$&&$\checkmark$&$\checkmark$&$\checkmark$&31.1&	53.7&28.9&24.9&40.3&49.0&38.1&14.6&41.9&43.8&15.3&	7.2&27.9&75.5&57.3&41.8&6.7&23.3&48.5&44.1&	35.7\\
\bottomrule[1.5pt]
\end{tabular}}
\end{table*}

\begin{table*}[h]

\caption{Results on adaptation from PASCAL VOC to Clipart Dataset with VGG. Average precision (\%) is evaluated on target images.}
\label{tbl:clipart_vgg}
  \centering
\scalebox{0.85}{
  \tabcolsep=1.5pt
  \centering
\begin{tabular}{c|ccc|ccccccccccccccccccccc}
\toprule[1.5pt]
%  & \multicolumn{20}{c}{AP for each class} & \\
%                \cmidrule(r){2-21}  
Method & G &  CTX &L&aero & bcycle & bird & boat & bottle & bus  & car  & cat  & chair & cow  & table & dog  & hrs & bike & prsn & plnt & sheep & sofa & train & tv & MAP  \\\hline
Faster RCNN&&&&15.7&31.9&22.4&8.2&38.8&59.4&17.8&6.6&37.0&5.7	&12.7&7.2&17.4&49.0&	36.0&32.1&11.2&2.9&29.8&28.4&23.5 \\
BDC-Faster &$\checkmark$&&&11.0	&40.9&12.2&10.1&28.9&29.2&28.0&6.0&	23.5&8.8&13.1&6.5	&22.4&45.6&46.9&35.9&9.7&9.3&18.9&20.9&	21.4 \\\hline
\multirow{2}{*}{Proposed} &$\checkmark$&&&18.6 & 45.0 & 22.2 & 23.2 & 23.9 & 21.1 & 28.6 & 5.2  & 31.8 & 39.1 & 19.7 & 0.9  & 25.2 & 56.1 & 54.3 & 36.1 & 27.8 & 6.8  & 32.4 & 40.4 & 27.9 \\
 &$\checkmark$&$\checkmark$&$\checkmark$&16.0 & 53.2 & 27.5 & 21.6 & 32.0 & 48.4 & 32.4 & 12.2 & 32.5 & 27.3 & 12.3 & 13.1 & 24.3 & 62.4 & 55.5 & 41.2 & 21.0 & 13.2 & 37.8 & 46.1 & 31.5\\
\bottomrule[1.5pt]
\end{tabular}}
\end{table*}

\begin{table*}[t]
\caption{Results on adaptation from PASCAL VOC to WaterColor Dataset ($\%$). Left: AP evaluated on PASCAL VOC. Right: AP evaluated on WaterColor. AP on adaptation from PASCAL VOC to WaterColor ($\%$). The definition of G, I, CTX, L is the same as defined in the main paper.  Faster RCNN* indicates Faster RCNN trained by source images and translated source images by CycleGAN.}
 \label{tbl:water_all2}
 \centering
 \scalebox{0.8}{
 \tabcolsep=1.5pt
\begin{tabular}{l|ccccc|ccccccc|ccccccc}
\toprule[1.5pt]
	&\multicolumn{5}{c|}{}  &   \multicolumn{7}{c}{AP on a source domain}& \multicolumn{7}{c}{AP on a target domain}\\%\cmidrule(r){6-19}
Method   & G & I & CTX & L &P & bike & bird & car  & cat  & dog  & prsn & MAP& bike & bird & car  & cat  & dog  & prsn & MAP  \\\hline
Faster RCNN &&&&& &82.1 & 82.3 & 86.5 & 89.3 & 85.6 & 84.3 & 85.0&68.8&46.8 & 37.2 & 32.7 & 21.3 & 60.7   & 44.6\\
{Faster RCNN}* &&&&&$\checkmark$ &79.7 & 82.7 & 86   & 88.9 & 85   & 82.1 & 84.1&83.3 & 52.7 & 45.3 & 33.1 & 28.8 & 64   & 51.2\\
BDC-Faster&$\checkmark$&&& &&80.9 & 82.6 & 86.4 & 87.9 & 82.7 & 83.3 & 84.0&68.6 & 48.3 & 47.2 & 26.5 & 21.7 & 60.5   & 45.5 \\
DA-Faster&$\checkmark$&$\checkmark$&&& &75.7 & 84.4 & 84.5 & 88.5 & 83.6 & 81.6 & 83.1 &75.2 & 40.6 & 48.0 & 31.5 & 20.6 & 60.0   & 46.0 \\\hline
\multirow{5}{*}{Proposed}  &$\checkmark$&&& &&79.4 & 84.6 & 85.8 & 89.6 & 85.7 & 84.1 & 84.9&66.4 & 53.7 & 43.8 & 37.9 & 31.9 & 65.3   & 49.8 \\
&&&& $\checkmark$&&80.0&82.0&84.7&86.8&83.6&81.3&83.1&79.4&54.8&47.2&37.1&31.5&62.4&52.1\\
 &$\checkmark$&& $\checkmark$&&&79.5 & 84.6 & 86.1 & 89.2 & 84.5 & 82.8 & 84.4& 71.3 & 52.0 & 46.6 & 36.2 & 29.2 & 67.3   & 50.4\\
 &$\checkmark$&& $\checkmark$& $\checkmark$&&79.8 & 87.4 & 85.5 & 88.1 & 84.5 & 84.0   & 84.9&82.3	& \bf{55.9}&46.5&32.7&35.5&66.7&53.3\\
 &$\checkmark$&& $\checkmark$& $\checkmark$& $\checkmark$& 78.9&83.1&84.2&87.4&85.6&82.8&83.7&\bf{90.5}&54.8&\bf{49.4}&\bf{38.6}&\bf{38.8}&\bf{67.9}&\bf{56.7}\\\hline
 Oracle&&&&&&82.1 & 82.3 & 86.5 & 89.3 & 85.6 & 84.3 & 85.0&83.6&59.4&50.7&43.7&39.5&74.5&58.6\\
\bottomrule[1.5pt]
\end{tabular}}
\end{table*}

\begin{table*}[h]
  \caption{Results on adaptation from Cityscape to FoggyCityscape Dataset ($\%$). The performance is evaluated on Cityscape.}
 \label{tbl:ap_foggy2}
 \centering
 \scalebox{0.8}{
 \tabcolsep=1.5pt
\begin{tabular}{l|cccc|cccccccccc}
\toprule[1.5pt]
%&  && 	& \multicolumn{8}{c}{AP for each class} & \\ \cmidrule(r){2-12}
	&\multicolumn{4}{c|}{}  &  \multicolumn{9}{c}{AP on a source domain}\\
Method& G & I & CTX & L & bus  &bcycl & car  & mcycl & prsn & rider & train & truck & MAP  \\\hline
Faster RCNN& &&&&55.6&39.0&52.3&38.8&33.7&47.7&39.1&33.1&42.4\\
NDC-Faster&$\checkmark$  &&& &56.4&38.2&	52.7&36.5&33.6&49.3&41.9&32.0&42.6 \\
DA-Faster&$\checkmark$  &$\checkmark$&&&58.3&38.8&52.6&42.5&35.1&47.6&44.1&34.2&44.2\\\hline
\multirow{3}{*}{Proposed} &$\checkmark$  &&&&62.6&37.9&52.2&35.1&35.0&48.5&47.7&34.9&44.2 \\
 &$\checkmark$  &&$\checkmark$ &&57.0&39.3&52.3&39.9&33.6&48.3&41.6&36.0&43.5\\
 &$\checkmark$  &&$\checkmark$ &$\checkmark$&57.9&39.4&52.6&39.4&35.2&48.3&47.7&37.4&44.7\\
\bottomrule[1.5pt]
\end{tabular}}
\end{table*}

\begin{table}[t!]
 \caption{Results on adaptation from Sim10k to Cityscape Dataset ($\%$). Average precision is evaluated on target images. Faster RCNN* indicates Faster RCNN trained by source images and translated source images by CycleGAN.}
 \label{tbl:ap_sim10k2}
 \centering
 \scalebox{0.75}{
\begin{tabular}{l|ccccc|c}
\toprule[1.5pt]
%&  && 	& \multicolumn{8}{c}{AP for each class} & \\ \cmidrule(r){2-12}
Method& G & I & CTX & L &P & AP on Car  \\\hline
Faster RCNN& &&&&&  34.6  \\
{Faster RCNN}*&  &&&&$\checkmark$& 40.0\\
BDC-Faster&$\checkmark$  &&&&&  31.8 \\
DA-Faster&$\checkmark$  &$\checkmark$&&&& 34.2\\\hline
\multirow{6}{*}{Proposed (FL)}&$\checkmark$  &&&&&   36.4 \\
&  &&&$\checkmark$&& 40.2\\
&$\checkmark$  &&$\checkmark$ &&&  38.2 \\
%&$\checkmark$  &&$\checkmark$ &&&   (38.3)* \\
&$\checkmark$  &&$\checkmark$&$\checkmark$ &&  40.1 \\
%Proposed & &&$\checkmark$&$\checkmark$ &&  39.8\\
%&&&&$\checkmark$&&  40.4 \\
%Proposed &$\checkmark$  &&&&$\checkmark$&  40.3 \\
 &$\checkmark$  &&$\checkmark$&&$\checkmark$&  41.5 \\
 &$\checkmark$  &&$\checkmark$&$\checkmark$&$\checkmark$&  40.7 \\\hline
 \multicolumn{7}{c}{Proposed Method with different parameters} \\\hline
FL ($\gamma=3$) &$\checkmark$  &&$\checkmark$ &&&  42.3 \\
Oracle &&&& &&  53.1 \\
\bottomrule[1.5pt]
\end{tabular}}
\end{table}

\end{document}